%% file: bare_jrnl_new_sample4.tex
\documentclass[lettersize,journal]{IEEEtran}
\usepackage{amsmath,amsfonts}
\usepackage{amssymb}
\usepackage{algorithmic}
\usepackage{algorithm}
\usepackage{array}
\usepackage[caption=false,font=normalsize,labelfont=sf,textfont=sf]{subfig}
\usepackage{textcomp}
\usepackage{stfloats}
\usepackage{url}
\usepackage{verbatim}
\usepackage{graphicx}
\usepackage{cite}
\usepackage{mathtools}
\usepackage{cleveref}
\usepackage{lipsum}
\usepackage{nicefrac}
\usepackage{makecell}
\usepackage{xcolor}
\usepackage{microtype}
\usepackage{booktabs}
\usepackage{multirow}
\usepackage{hhline}
\newcommand{\ourmodel}{GeoDTR\text{+}}

\newcommand{\Rone}{R@$1$}
\newcommand{\Rfive}{R@$5$}
\newcommand{\Rten}{R@$10$}
\newcommand{\Ronep}{R@$1\%$}
\newcommand{\chsg}{CHSG}

\newcommand{\B}[1]{
    {\color{magenta} \bf #1}
}
\newcommand{\SB}[1]{
    {\color{cyan} \bf #1}
}

\newcommand{\ro}[1]{\textcolor{black}{#1}} 
\newcommand{\rt}[1]{\textcolor{black}{#1}} 
\newcommand{\common}[1]{\textcolor{black}{#1}} 

\begin{document}

\title{GeoDTR\text{+}: Toward Generic Cross-View Geolocalization via Geometric Disentanglement}

\author{Xiaohan Zhang\textsuperscript{\dag}, Xingyu Li\textsuperscript{\dag}, Waqas Sultani, Chen Chen, Safwan Wshah\textsuperscript{\ddag}
\thanks{$\dag$ These authors contributed equally.}
\thanks{$\ddag$ Corresponding and senior author.}
\thanks{X.Zhang and S.Wshah are with Department of Computer Science and Vermont Complex Systems Center, University of Vermont, Burlington, USA}
\thanks{X. Li is with Lin Gang Laboratory, Shanghai, China}
\thanks{W. Sultani is with Intelligent Machine Lab, Information Technology University, Pakistan}
\thanks{C. Chen is with Center for Research in Computer Vision, University of Central Florida, USA}
}

\markboth{Journal of \LaTeX\ Class Files,~Vol.~14, No.~8, August~2021}%
{Shell \MakeLowercase{\textit{et al.}}: A Sample Article Using IEEEtran.cls for IEEE Journals}



\newpage

\maketitle

\input{text/abstract}

\input{text/introduction}
\input{text/related_work}

\input{text/methodology}
\input{text/experiments}

\input{text/discussion}
\input{text/conclusion}
 
%

\bibliographystyle{IEEEtran}

\bibliography{references}

\clearpage
\input{text/Appendix}

\clearpage

\section{Biography Section}

\begin{IEEEbiography}[{\includegraphics[width=1in,height=1.25in,clip,keepaspectratio]{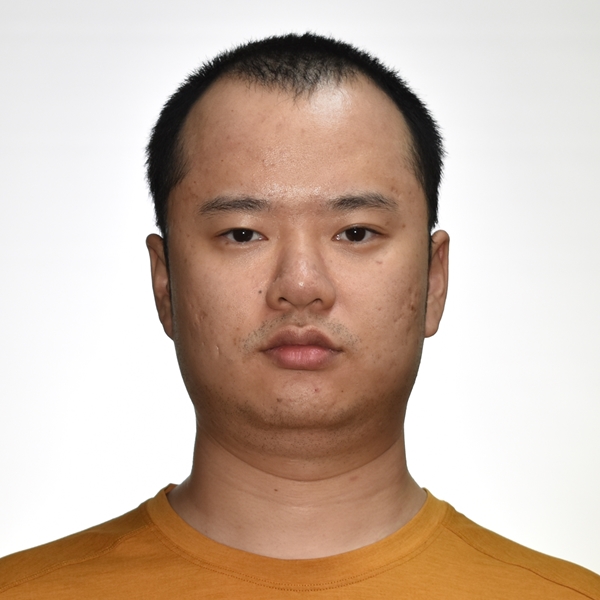}}]{Xiaohan Zhang}
is a third-year Ph.D. student at the University of Vermont advised by Dr.Safwan Wshah. His current research interest lies at the border of computer vision and remote sensing (e.g. visual geo-localization and segmentation/detection in aerial images). He is also interested in image synthesis and 3D reconstruction. Before joining the University of Vermont, he received his M.Sc. degree at the University of California, Santa Cruz in 2020, and he obtained his B.Sc. degree at Michigan State University in 2017.
\end{IEEEbiography}

\begin{IEEEbiography}[{\includegraphics[width=1in,height=1.25in,clip,keepaspectratio]{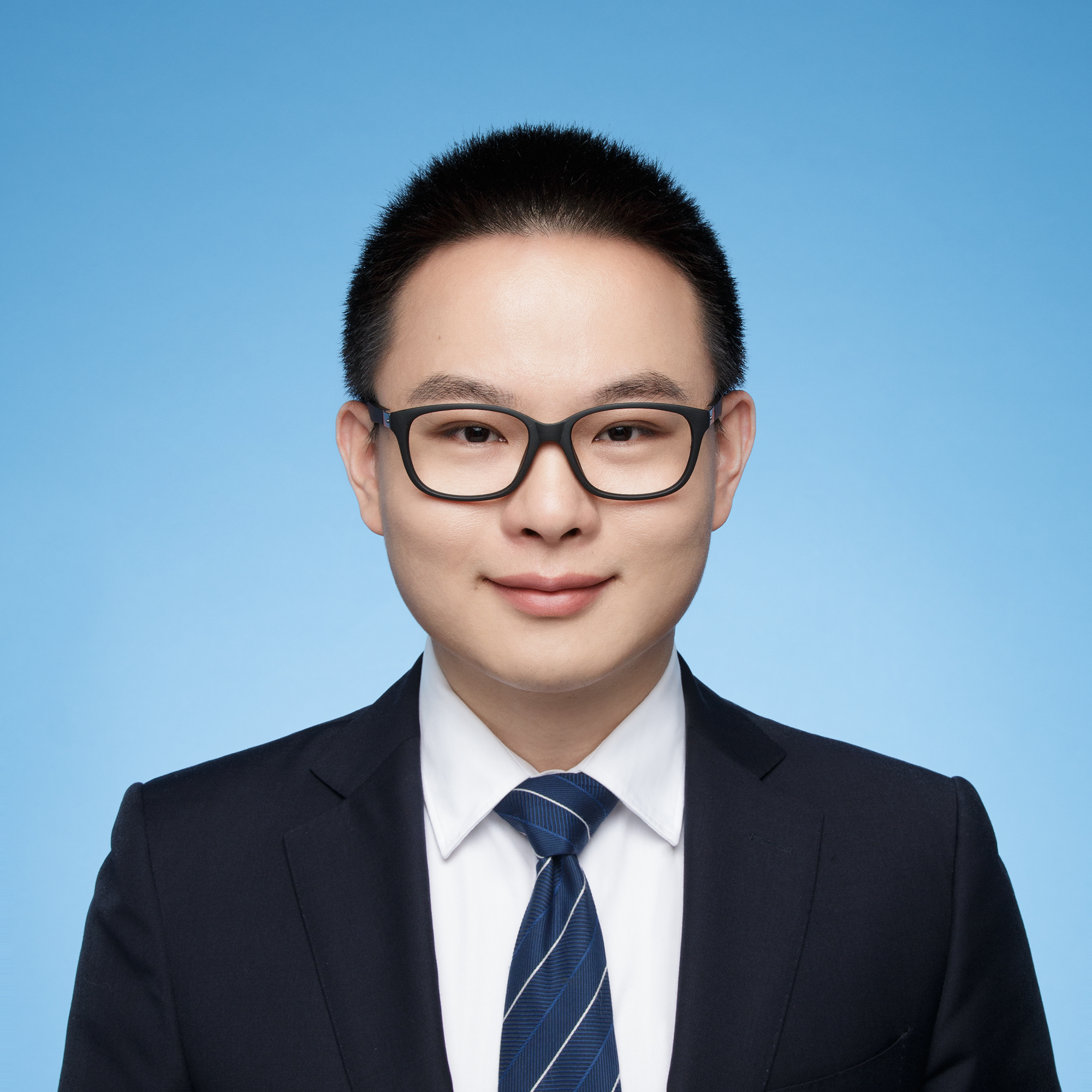}}]{Dr. Xingyu Li}
is currently an Assistant Researcher at Lin Gang Laboratory, Shanghai.
He received his M.Sc. degree in computer science from the University of California, Santa Cruz in 2020, and his Ph.D. degree in atomic and molecular physics from the University of Science and Technology of China, Hefei in 2017. He was a post-doctoral researcher with Shanghai Center for Brain Science and Brain-Inspired Technology during 2021.07-2023.07. His research interests include deep learning, cognitive neuroscience, and brain-inspired intelligence. 
\end{IEEEbiography}

\begin{IEEEbiography}[{\includegraphics[width=1in,height=1.25in,clip,keepaspectratio]{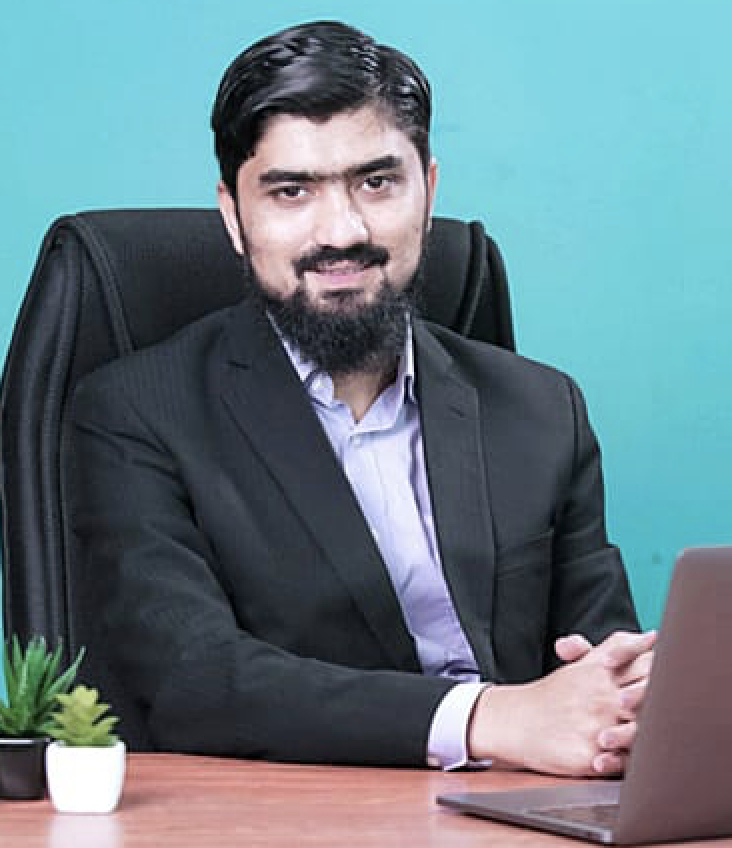}}]{Dr. Waqas  Sultani}
is an Assistant Professor at the Information  Technology University, Lahore. He received his Ph.D. from the 
Center for Research in Computer Vision at UCF. His research related to human action recognition, anomaly detection, small object detection, geolocalization, and medical imaging was published in the top venues including CVPR, ICRA, AAAI, WACV, CVIU, MICCAI, etc. He was awarded the Facebook-CV4GC research award in 2019 and a Google Research Scholar award in 2023 for projects related to medical imaging. He also served as an area chair for several conferences such as CVPR 2022-2024, and ACM-MM 2020-2021. 
\end{IEEEbiography}

\begin{IEEEbiography}[{\includegraphics[width=1in,height=1.25in,clip,keepaspectratio]{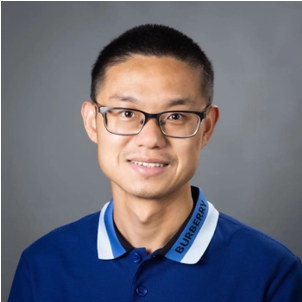}}]{Dr. Chen Chen}
is an Assistant Professor at the Center for Research in Computer Vision at UCF. He received his Ph.D. in Electrical Engineering from UT Dallas in 2016, receiving the David Daniel Fellowship (Best Doctoral Dissertation Award). His research interests include computer vision, efficient deep learning, and federated learning. He has been actively involved in several NSF and industry-sponsored research projects, focusing on efficient resource-aware machine vision algorithms and systems development for large-scale camera networks. He is an Associate Editor of IEEE Transactions on Circuits and Systems for Video Technology (T-CSVT), Journal of Real-Time Image Processing, and IEEE Journal on Miniaturization for Air and Space Systems. He also served as an area chair for several conferences such as ECCV’2022, CVPR’2022, ACM-MM 2019-2022, ICME 2021 and 2022. According to Google Scholar, he has 16K+ citations and an h-index of 63.
\end{IEEEbiography}

\begin{IEEEbiography}[{\includegraphics[width=1in,height=1.25in,clip,keepaspectratio]{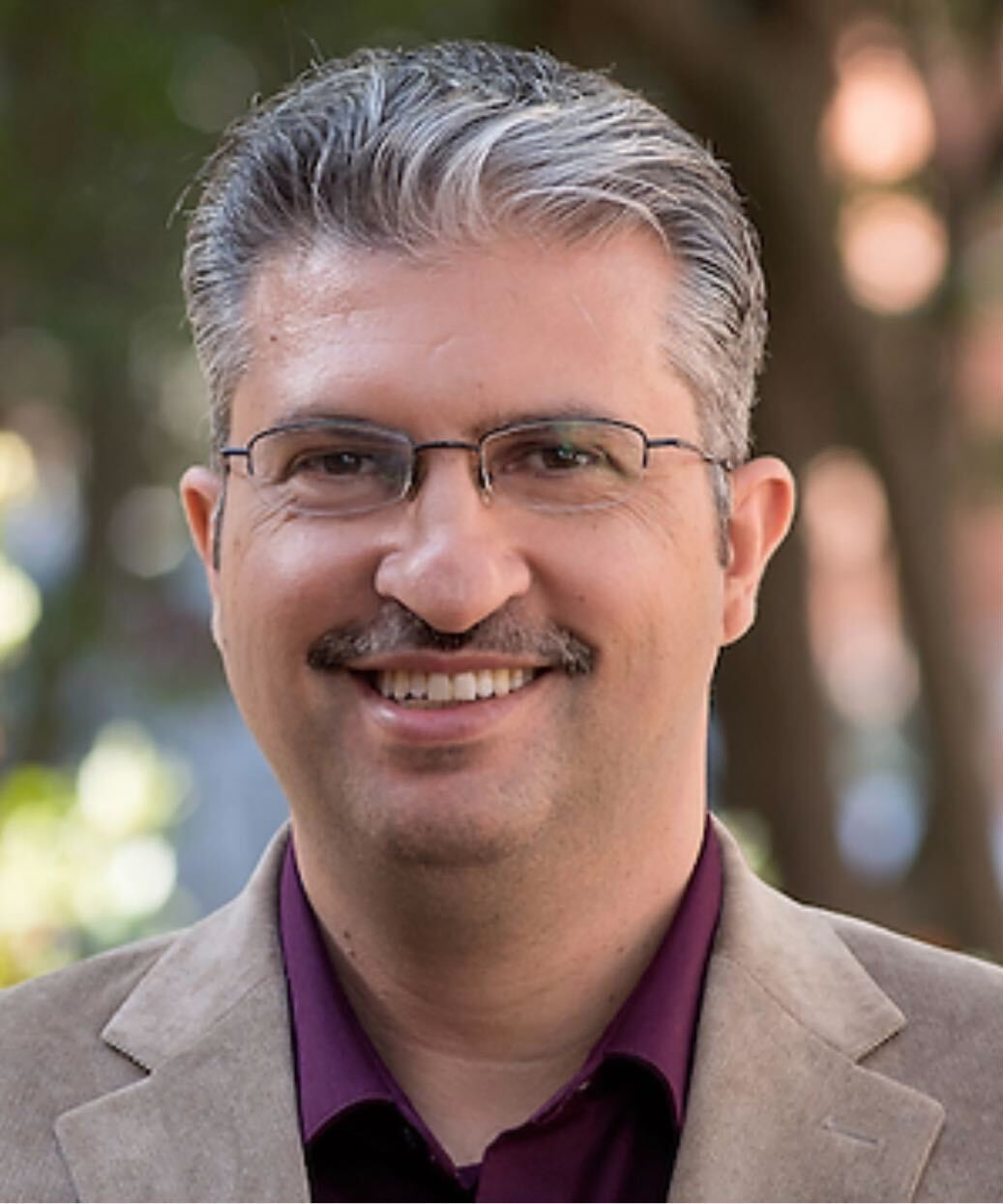}}]{Dr. Safwan Wshah}
is currently an Associate Professor in the Department of Computer Science at the University of Vermont. His research interests lie at the intersection of machine learning theory and application. His core area is object understanding and geo-localization from the ground, aerial, and satellite images. He also has broader interests in deep learning, computer vision, data analytics, and image processing. Dr. Wshah received his Ph.D. in Computer Science and Engineering from the University at Buffalo in 2012. Prior to joining the University of Vermont, Dr. Wshah worked for Xerox and PARC (Palo Alto Research Center)- Xerox company, where he was involved in several projects creating machine learning algorithms for different applications in healthcare, transportation, and education fields.
\end{IEEEbiography}

\end{document}

%% file: text/abstract.tex
\begin{abstract}
Cross-View Geo-Localization (CVGL) estimates the location of a ground image by matching it to a geo-tagged aerial image in a database. Recent works achieve outstanding progress on CVGL benchmarks. However, existing methods still suffer from poor performance in cross-area evaluation, in which the training and testing data are captured from completely distinct areas. We attribute this deficiency to the lack of ability to extract the geometric layout of visual features and models' overfitting to low-level details. 
Our preliminary work~\cite{GeoDTR} 
introduced a Geometric Layout Extractor (GLE) to capture the geometric layout from input features. However, the previous GLE does not fully exploit information in the input feature.
In this work, we propose \ourmodel{} with an enhanced GLE module that better models the correlations among visual features. 
To fully explore the LS techniques from our preliminary work, we further propose Contrastive Hard Samples Generation (\chsg{}) to facilitate model training. 
Extensive experiments show that \ourmodel{} achieves state-of-the-art (SOTA) results in cross-area evaluation on CVUSA~\cite{CVUSA}, CVACT~\cite{liu2019lending}, and VIGOR~\cite{Vigor} by a large margin ($16.44\%$, $22.71\%$, and $13.66\%$ without polar transformation) while keeping the same-area performance comparable to existing SOTA. 
Moreover, we provide detailed analyses of \ourmodel{}. Our code will be available at \textcolor{magenta}{\url{https://gitlab.com/vail-uvm/geodtr_plus}}.
\end{abstract}

\begin{IEEEkeywords}
Visual Geolocalization, Cross-view Geolocalization, Image Retrieval, Metric Learning
\end{IEEEkeywords}

%% file: text/introduction.tex
\section{Introduction}
\label{sec::introduction}

\begin{figure}
    \centering
    \includegraphics[width=0.48\textwidth]{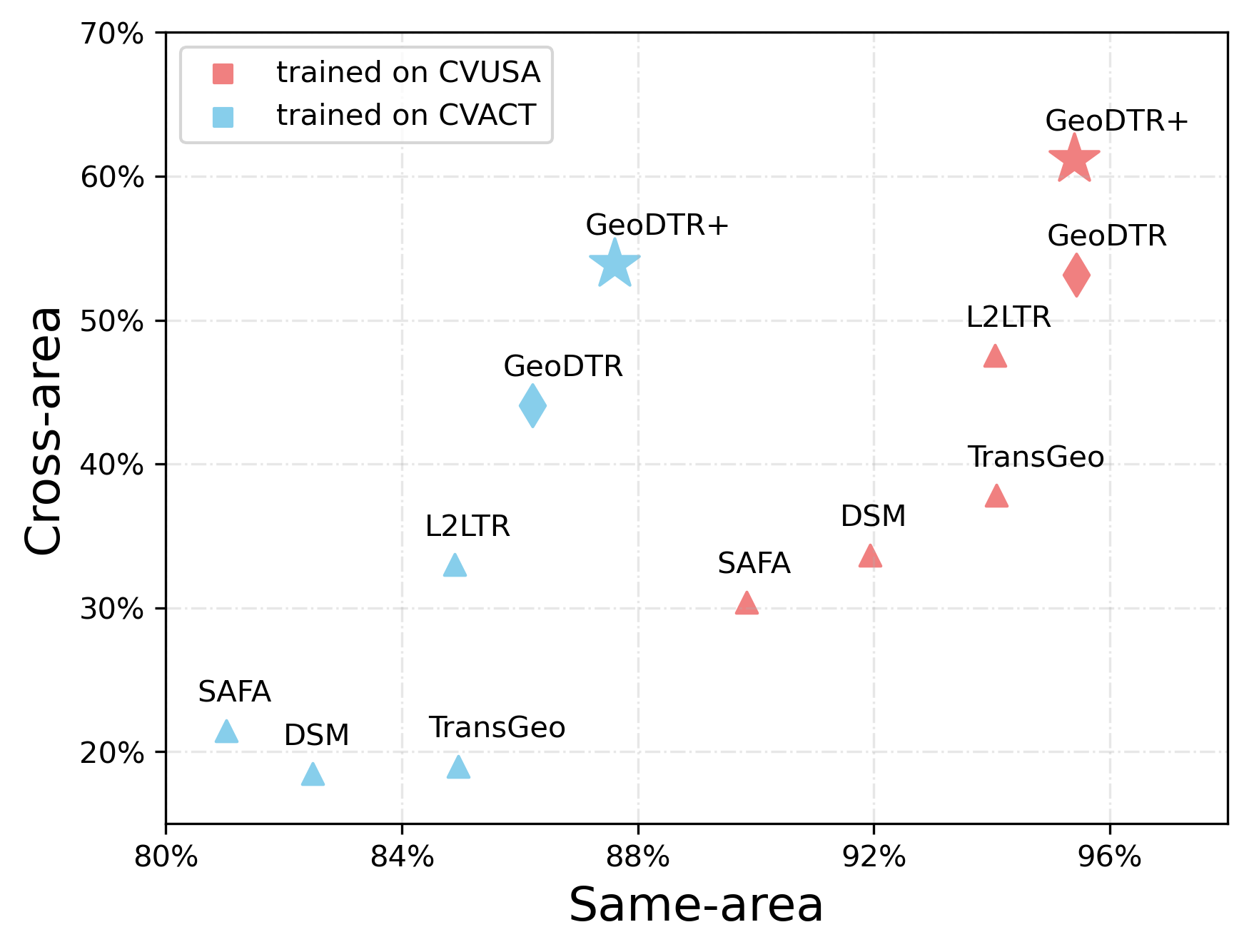}
    \caption{Comparison of \Rone{} accuracy between four recently published CVGL methods, including SAFA~\cite{SAFA}, DSM~\cite{DSM}, TransGeo~\cite{transgeo}, L2LTR~\cite{l2ltr}, GeoDTR~\cite{GeoDTR}, and our proposed \ourmodel{} on same-area performance (x axis) and cross-area performance (y axis). Notice that our preliminary work GeoDTR achieves the SOTA same-area and cross-area performance. In this work, built upon the GeoDTR, the proposed \ourmodel{} further improves more, especially on cross-area performance.}
    \label{fig:teaser}
\end{figure}

Estimating the location of a ground image from a database of geo-tagged aerial images, named ``Cross-View Geo-Localization (CVGL)'', is one of the fundamental tasks that lies at the border of remote sensing and computer vision. The ground image is referred to as the query image and the geo-tagged aerial images are known as reference images. Different from same-view geo-localization which utilizes a geo-tagged ground images database for referencing, CVGL takes aerial images as reference data that is more accessible~\cite{survey}. CVGL facilitates various tasks such as autonomous driving~\cite {kim2017satellite}, unmanned aerial vehicle navigation~\cite{UAV1}, object localization~\cite{object_geolocalization}, and augmented reality~\cite{AR1} by providing more accurate location estimation under certain environments. Existing approaches typically treat CVGL as an image retrieval problem~\cite{survey}. This process extracts features from both ground and aerial images by either handcrafted descriptors or learned neural networks. The estimated location is the \ro{one which shares the most similar latent features to the query ground image~\cite{SAFA,DSM,l2ltr,cvmnet, Vo, CVUSA,comingDown,featureTransport,liu2019lending,wang2021each,zheng2020university,Vigor,transgeo,SAIG}.} 
To achieve this goal, these methods normally train a model to push the corresponding aerial image and ground image pairs (also known as aerial-ground pairs) closer in latent space and push the unmatched pairs further away from each other.

CVGL is considered an extremely challenging problem because of: (i) the drastic difference in the viewing angles between ground and aerial images, (ii) the variance in capturing time of both ground and aerial images, and (iii) very different resolutions between ground and aerial images.
Addressing such challenges requires a comprehensive and profound understanding of the image content and the spatial configuration (i.e. the relative locations between each landmark in an image). Most existing methods~\cite{SAFA,cvmnet,lin2015learning,featureTransport,hardTriplet,Vo} pair the query ground image and reference image by exploiting features extracted by Convolutional Neural Networks (CNNs). \ro{For example, Spatial-Aware Feature Aggregation (SAFA)~\cite{SAFA} proposed to explore the CNN extracted feature by a customized Spatial-aware Positional Embedding (SPE) module which explores the spatial configuration by fully connected layers. Such methods are limited in exploring the spatial configuration which is a global property (i.e. a highway spanning west to east on an aerial image) due to the capacity of fully connected layers that can hardly explore correlations explicitly.} Recently, transformer~\cite{attention} has been introduced in CVGL~\cite{l2ltr,transgeo,Zhang_2023_WACV} to capture the global contextual information. Nevertheless, in these methods,
the spatial configuration is unavoidably entangled with low-level features because the multi-head attention mechanism implicitly models those correlations.

Typically, CVGL models are expected to generalize on unseen data with minimal supervision, for example, estimating the locations of autonomous vehicles or smartphones at any geospatial point. Furthermore, popular CVGL methods heavily rely on a certain area with high-quality aerial-ground image pairs for effective training, which is not always accessible at other locations. Consequently, there is a pressing need to develop generalizable CVGL algorithms capable of estimating the location of ground query data from any geographic area which is also known as cross-area evaluation (i.e. training and testing on two distinct geographic areas). Despite the evaluation of existing CVGL algorithms on datasets where the training and testing data originate from both the same geographical area (same-area) and different geographical locations (cross-area), \ro{the current emphasis within the CVGL field is primarily placed on improving same-area performance~\cite{SAFA,cvmnet,DSM,transgeo,SAIG}. However, there is a lack of a method for generalizing the model on cross-area benchmarks.} This results in a significant performance gap between same-area and cross-area performance and a lack of focus on achieving satisfactory cross-area performance. For example, as shown in \Cref{fig:teaser}, on CVUSA~\cite{CVUSA} dataset, SAFA~\cite{SAFA}, L2LTR~\cite{l2ltr}, DSM~\cite{DSM}, and TransGeo~\cite{transgeo} achieve around $90\%$ \Rone{} accuracy on same-area evaluation but only achieve $30\%$ to $40\%$ \Rone{} accuracy on cross-area performance. Furthermore, on CVACT~\cite{liu2019lending} dataset, these methods can only achieve $20\%$ to $30\%$ \Rone{} accuracy on cross-area performance while achieving more than $80\%$ \Rone{} accuracy on same-area performance.

As demonstrated in \Cref{fig:teaser}, there is a notable gap between the same-area and the cross-area performance of these algorithms. This paper seeks to answer the question - \textit{how to generalize cross-view geo-localization algorithms with minimal supervision?}

A preliminary version of this work has been published at the 37th AAAI conference on artificial intelligence~\cite{GeoDTR}. In that work, we proposed a Geometric Layout Extractor (GLE) module to capture spatial configuration, 
as well as Layout simulation and Semantic augmentation (LS) techniques to diversify the training data. Moreover, 
a counterfactual (CF) learning schema was introduced to provide additional supervision for the geometric layout extractor module. However, the proposed GLE module only explores geometric correlations within a learned subspace, which may result in a loss of information. Additionally, the effectiveness of the proposed LS techniques in enhancing cross-area performance remains limited, as they were only utilized in a data augmentation manner in the preliminary work. To address the above-mentioned limitations, this work extends the previous study in the following three aspects.
First, we improve the design of the geometric layout extractor module to better capture the spatial configurations in both aerial and ground images (\Cref{sec::GLE}). 
Second, in our previous work, LS techniques are adopted as a special data augmentation that implicitly introduces inter-batch contrastive signals for layout and semantic features. Inspired by the hard sample mining strategy in cross-view geo-localization~\cite{Vigor}, we extend LS techniques to explicitly include intra-batch contrastive signals in the current work (\Cref{sec::chsg}).
In this way, our \ourmodel{} is expected to distinguish these ``generated hard'' samples in a single batch. We named this process the ``Contrastive Hard Samples Generation'' (\chsg) procedure. By integrating \chsg{} with the novel geometric layout extractor, \ourmodel{} is able to learn better latent representations explicitly from generated hard samples within a single batch. This results in a substantial improvement in cross-area performance compared to the current state-of-the-art (SOTA) methods. Furthermore, we provide more comprehensive benchmark results of \ourmodel{} with other CVGL methods on more challenging datasets~\cite{Vigor} and also provide additional analyses (\Cref{sec::experiment}) and visualization examples. 
Our contribution to this paper can be summarized as threefold,
\begin{itemize}
    \item We present \textbf{Geo}metric \textbf{D}escriptor \textbf{TR}ansformer plus (\ourmodel{}), a novel CVGL model that is built upon our preliminary GeoDTR~\cite{GeoDTR} model. \ourmodel{} can effectively model the spatial configuration of both ground and aerial images through the proposed novel geometric layout extraction mechanism. 
    \item We enhance our approach by incorporating LS techniques with the Contrastive Hard Samples Generation (\chsg{}) process. This combination provides our model with improved guidance, enabling it to capture geometric layout information more effectively rather than overfitting to low-level details. CHSG plays a crucial role in this enhancement by allowing the model to explicitly distinguish and generate hard samples within a single batch. As a result, our model achieves significantly improved cross-area performance.

    \item Extensive experiments demonstrate that \ourmodel{} trained with \chsg{} surpasses the current state-of-the-art cross-area performance by a considerable margin on popular datasets CVUSA~\cite{CVUSA}, CVACT~\cite{liu2019lending}, and VIGOR~\cite{Vigor} (i.e. $16.44\%$, $22.71\%$, and $13.66\%$ without polar transformation on each of the three datasets respectively) without compromising the same-area performance. 
\end{itemize}

%% file: text/related_work.tex
\section{related work}
\label{sec::related_works}
\subsection{Cross-view Geo-localization}
\label{sec::related_works_CVGL}

\subsubsection{Feature-based CVGL}
\ro{Feature-based geo-localization methods extract both aerial and ground latent representations from local information using hand-crafted features or deep learning models without any geometric priors~\cite{lin2013cross, lin2015learning, CVUSA, survey}.} Existing works studied different aggregation strategy~\cite{cvmnet}, training paradigm~\cite{Vo}, loss functions (i.e. HER~\cite{hardTriplet} and SEH~\cite{SEH}) and feature transformation (i.e. feature fusion~\cite{bridging} and CVFT~\cite{featureTransport}).
The above-mentioned feature-based methods did not fully explore the effectiveness of spatial information due to the locality of CNN which lacks the ability to explore global correlations. By leveraging the ability to capture global contextual information of the transformer, our \ourmodel{} learns the geometric correspondence between ground images and aerial images through a transformer-based sub-module which results in a better performance.

\subsubsection{Geometry-based CVGL}
\ro{Recently, learning to match the geometric correspondence between aerial and street views has become a hot topic in cross-view geo-localization. Specifically, these methods leverage geometric priors such as polar transformation, heading, and orientation information or learning to capture non-local correlations to better predict the similarities in aerial and ground latent featuers~\cite{survey}.} Liu et al.~\cite{liu2019lending} proposed a model with encoded camera orientation in aerial and ground images.
Shi et al., in \cite{SAFA} proposed SAFA which aggregates features through its learned geometric correspondence from ground images and polar transformed aerial images. Later, the same author proposed Dynamic Similarity Matching (DSM)~\cite{DSM} to geo-localize limited field-of-view ground images by a sliding-window-like algorithm. CDE~\cite{comingDown} combined GAN~\cite{gan} and SAFA~\cite{SAFA} to learn cross-view geo-localization and ground image generation simultaneously. Despite the remarkable performance achieved by these geometric-based methods, they are limited by the nature of CNNs which explore the local correlation among pixels. 

Recent research~\cite{TGCNN,MGTL,transgeo} explores multi-head attention mechanism~\cite{attention} to capture non-local correlations in the images. \rt{TGCNN~\cite{TGCNN} adopted transformers to explore global and local correlations simultaneously. MGTL~\cite{MGTL} leveraged mutual interaction between aerial and ground images from a customized transformer mudle.} L2LTR~\cite{l2ltr} studied a hybrid ViT-based~\cite{ViT} method while TransGeo~\cite{transgeo} proposed the first transformer-only model for cross-view geolocalization. 
Despite the ability to explore spatial correlation, the above-mentioned methods do not process this global contextual information separately from others, such as the low-level features.
In this sense, they are of a single-pathway nature.

Differently, our proposed \ourmodel{} employs a two-pathway design in which one pathway solely engages in the explicit modeling of global contextual information. 
The quality of this global contextual information is further strengthened by our counterfactual (CF) learning schema and Contrastive Hard Samples Generation (\chsg{}). Benefitting from these designs, \ourmodel{} does not solely rely on the polar transformed aerial view which bridges the domain gap between aerial view and ground view in the pixel space. Moreover, \ourmodel{} has fewer trainable parameters than L2LTR~\cite{l2ltr} and furthermore, it does not require the 2-stage training paradigm as proposed in TransGeo~\cite{transgeo}.

\subsubsection{Data Augmentation in CVGL}
Data augmentation is widely used in computer vision. Nonetheless, its application in cross-view geo-localization is not fully explored due to the fragility of the spatial correspondence between aerial and ground images which can be easily disrupted by even minor perturbations. Some existing methods attempt to address this issue by randomly rotating or shifting one view while fixing the other one~\cite{liu2019lending, Rodrigues_2022_WACV, Vo, hardTriplet}. Alternatively, \cite{rodrigues2021these} randomly blackout ground objects according to their segmentation from street images. 
In our preliminary paper, we propose LS techniques that \textit{maintain} geometric correspondence between images of the two views while varying geometric layout and visual features during the training phase. Our extensive experiments demonstrate that LS can significantly improve the performance on cross-area datasets not only for \ourmodel{} but also can be universally applied to other existing methods.

\subsubsection{Sample Mining in CVGL}
Many existing CVGL methods explore data mining techniques to improve performance. In-batch mining methods~\cite{hardTriplet,SEH} leverage loss functions such as HER and the SEH to emphasize hard samples in a single batch. Global mining methods~\cite{revisiting,Vigor} maintain a global mining pool to construct training batches with hard samples drawn from the entire training dataset. However, in-batch mining methods are limited by the lack of sample diversity within a training batch, while global mining strategies require additional memory and computational resources to maintain the mining pool. In this paper, we propose the Contrastive Hard Sample Generation (\chsg{}) process, which leverages the merit of LS techniques to generate ``hard samples'' (aerial-ground pairs) with the corresponding geometric layout in a single batch without requiring additional computational resources. \chsg{} supervises the model to distinguish those `hard samples' by learning to extract the spatial configuration and avoid overfitting to low-level details. Thus, it alleviates the problem that most of the negative samples contribute small losses, and the convergence becomes slow when training progress approaches the end. \chsg{} only operates in forward pass. Thus it is as efficient as the in-batch mining methods mentioned above. Furthermore, \chsg{} does not require additional memory and computational resources to maintain the mining pool.

\subsection{Counterfactual Learning}
The idea of counterfactual in causal inference~\cite{Pearl_Causality} has been successfully applied in several research areas such as explainable artificial intelligence ~\cite{explain_cf}, visual question answering~\cite{VQA_cf}, physics simulation~\cite{cophy_cf}, and reinforcement learning~\cite{reinforce_cf}. Inspired by these recent successes in counterfactual causal inference, in our preliminary paper~\cite{GeoDTR}, we propose a novel distance-based counterfactual (CF) learning schema that strengthens the quality of learned geometric descriptors for our \ourmodel{} and keeps it away from an obvious `wrong' solution. Our preliminary paper demonstrates that the proposed CF learning schema improves the performance of the model in both same-area and cross-area evaluation benchmarks. In this paper, we keep this CF learning schema in the \ourmodel{} to improve the quality of descriptors from the proposed novel geometric layout extractor.

%% file: text/methodology.tex
\section{methodology}
\label{sec::methodology}

\begin{figure*}[!h]
    \centering
    \includegraphics[page=6,width=1.0\textwidth, trim={0cm 60cm 0cm 4cm}, clip]{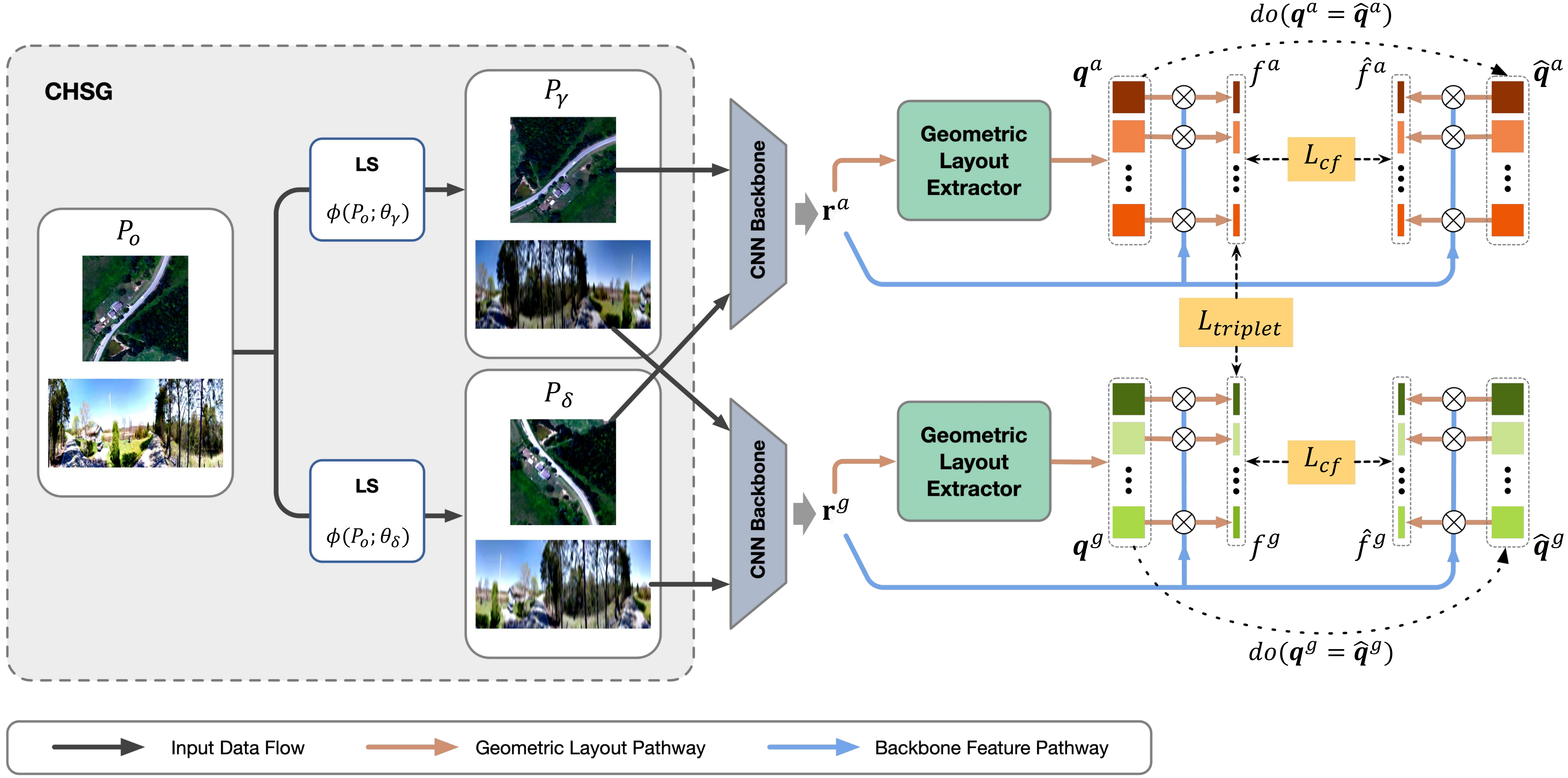}
    \caption{The overview pipeline of our proposed model \ourmodel{}. The Contrastive Hard Sample Generation (\chsg{}) first samples an aerial-ground pair $P_{o}$ from the training dataset and generate hard samples $P_{\gamma}$ and $P_{\delta}$. The proposed Geometric Layout Extractor (GLE) predicts the layout descriptors $\mathbf{q}^{a(g)}$ from the raw feature $\mathbf{r}^{a(g)}$. The predicted latent representation $f^{a(g)}$ is obtained from the frobenius product between $\mathbf{r}^{a(g)}$ and $\mathbf{q}^{a(g)}$. The proposed counterfactual learning provides an auxiliary supervision signal to train the model.}
    \label{fig:model}
\end{figure*}

\begin{figure}[!h]
    \centering
    \includegraphics[page=8,width=0.45\textwidth, trim={1cm 40cm 41cm 0cm}, clip]{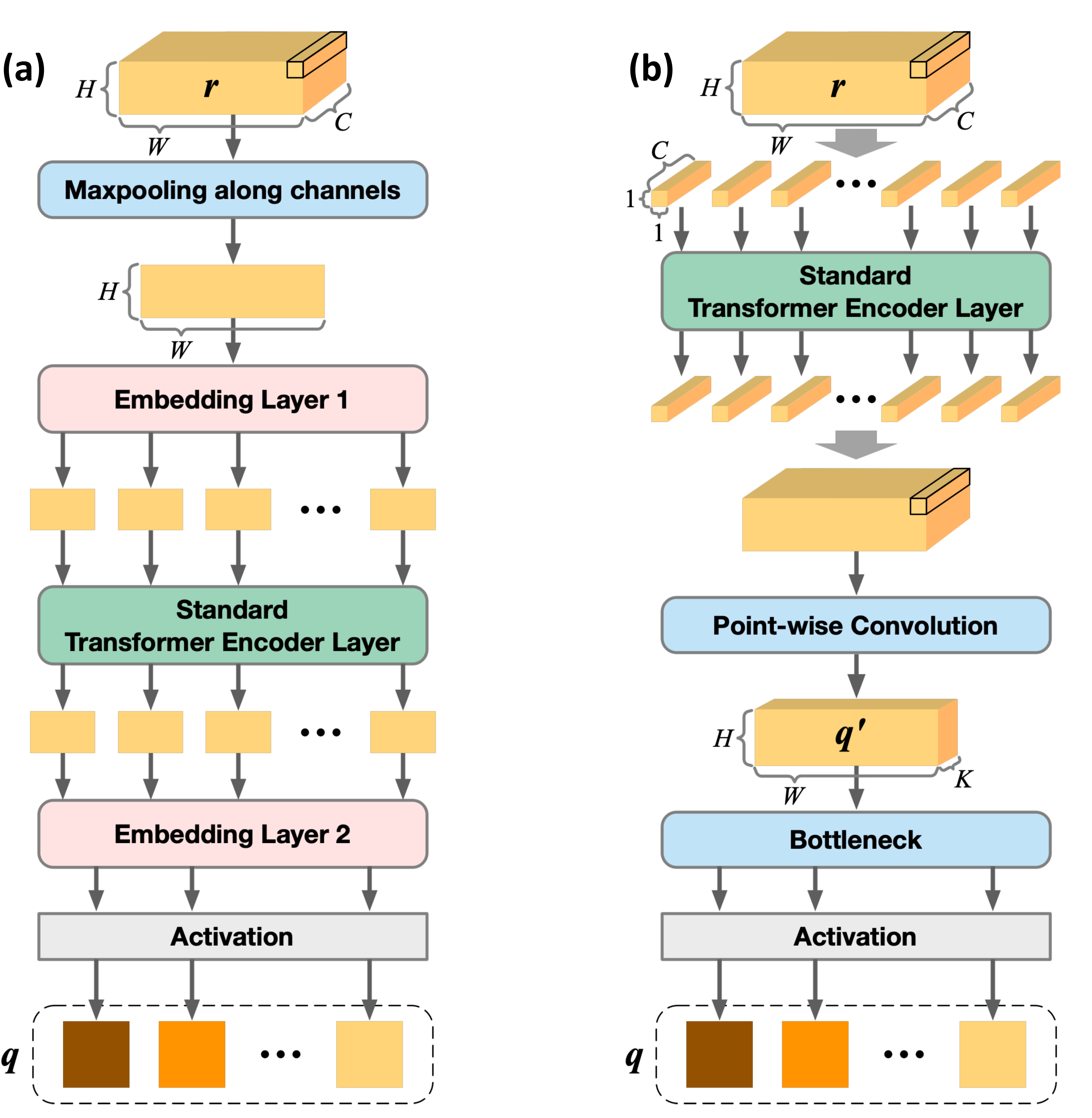}
    \caption{Comparison between the previous GLE and the proposed GLE. (a) is the GLE from our previous GeoDTR~\cite{GeoDTR}. (b) is the enhanced GLE for \ourmodel{}.}
    \label{fig:GLE}
\end{figure}

\subsection{Problem Formulation\label{sec:problemformulation}}


\ro{Consider a set of ground images $\{I^g_i\}, i=1,\dots,M$ and a set of aerial images $\{I^a_i\}, i=1,\dots,N$ where superscripts $g$ and $a$ are abbreviations for ground and aerial, respectively. $M$ and $N$ are the number of aerial and ground images.}

In the cross-view geo-localization task, given a query ground image $I^g_{y}$
with index $y$, 
one searches for the best-matching reference aerial image $I^a_b$ with $b\in\{1,\dots,N\}$.

For the sake of a feasible comparison between a ground image and an aerial image, we seek discriminative latent representations $f^g$ and $f^a$ for the images. These representations are expected to 
capture the dramatic view-change as well as the abundant low-level details, such as textual patterns.
Then the image retrieval task can be made explicit as
\begin{equation}\label{eq:task}
    b = \underset{i\in\{1, \dots, N\}}{\arg\min}\, d(f^g_{y}, f^a_i),
\end{equation}
where $d(\cdot,\cdot)$ denotes the $L_2$ distance. For the compactness in symbols, we will use superscript $v$ for cases that apply to both ground ($g$) and aerial ($a$) views. We adopt this convention throughout the paper.

\subsection{Geometric Layout Modulated Representations}
To generate high-quality latent representations for cross-view geo-localization, we emphasize the spatial configurations of visual features as well as low-level features. The spatial configuration reflects not only the positions but also the global contextual information among visual features in an image. 
One could expect such geometric information to be stable during the change of views. Meanwhile, the low-level features such as color and texture, help to identify visual features across different views. 

Specifically, we propose the following decomposition of the latent representation
\begin{equation}
    f^v = \mathbf{q}^v \circ \mathbf{r}^v.
    \label{eq:main}
\end{equation}
$\mathbf{q}^v=\{q^v_{m}\}_{{m}=1,\dots,K}$ is the set of $K$ geometric layout descriptors that summarize the spatial configuration of visual features, and $\mathbf{r}^v=\{r^v_j\}_{j=1,\dots,C}$ denotes the raw latent representations of $C$ channels that is generated by any backbone encoder. Both $q^v_{m}$ and $r^v_j$ are \ro{matrices} in $\mathbb{R}^{H\times W}$ with $H$ and $W$ being the height and width of the raw latent representations, respectively.
The modulation operation $\mathbf{q}^v \circ \mathbf{r}^v$ expands as 
\begin{equation}
\label{eq::f_product}
\left( \langle q^v_1, r^v_1 \rangle,\dots, \langle q^v_1, r^v_C \rangle, \dots,  \langle q^v_K, r^v_1 \rangle,\dots, \langle q^v_K, r^v_C \rangle \right),
\end{equation}
where $\langle p^v_{m}, r^v_j \rangle$ denotes the Frobenius inner product of $p^v_{m}$ and $r^v_j$.
In this sense, the resulting $f^v \in \mathbb{R}^{CK}$ are referred to as the \emph{geometric layout modulated representations} and will be used in~\Cref{eq:task} to retrieve the best-matching aerial images.
Our model design closely follows the above decomposition.

\subsection{\ourmodel{} Model}
\label{sec::model}

\subsubsection{Model Overview}
\ourmodel{} (see~\Cref{fig:model}) is a siamese neural network including two branches for the ground and the aerial views, respectively.
Within a branch, there are two distinct processing pathways, i.e., the backbone feature pathway and the geometric layout pathway.
Furthermore, we introduce Contrastive Hard Samples Generation (\chsg{}) to construct training batches with hard aerial-ground pairs. 

In the backbone feature pathway, a CNN backbone encoder processes the input image to generate raw latent representations $\mathbf{r}^v$ where $v=g \text{ or } v=a$. Due to the nature of the CNN backbone, these representations carry the positional information as well as the low-level feature information.

The geometric layout pathway is devoted to exploring the global contextual information among visual features. 
This pathway includes a core sub-module called the Geometric Layout Extractor (GLE), 
which generates a set of geometric layout descriptors (GLDs) $\mathbf{q}^v$ based on the raw latent representations $\mathbf{r}^v$.
These descriptors will modulate $\mathbf{r}^v$, integrating the geometric layout information therein.
With a stand-alone treatment of the geometric layout, one avoids introducing undesired correlations among the low-level features from different visual features. In the following, we will describe the key components of \ourmodel{} in detail.

\subsubsection{Geometric Layout Extractor}
\label{sec::GLE}
Geometric Layout Extractor (GLE) mines the global contextual information, such as the relative locations of buildings and roads, among the visual features and thus produces GLDs containing these global layout patterns. Despite the change in appearance across views, the arrangement of visual features remains largely intact. Hence, integrating the geometric layout information into the latent representations $f^v$ would improve its discriminative power for cross-view geo-localization. Note that the geometric layout is a global property in the sense that it captures the spatial configuration of single/multiple visual features at different positions in the images. For example, a single visual feature can span across the image, such as the road. In our model, the GLDs plays the role that grasps the global correlation among visual features.

In our GeoDTR preliminary paper, the GLE therein (see~\Cref{fig:GLE}a) employs a max pooling layer that is applied along the channel dimension to produce a saliency map. Then GLDs are generated by a transformer module that explores correlations in sub-spaces projected from this saliency map. However, such a mechanism suffers from information loss during the saliency map generation, leading to sub-optimal performance.

In this paper, we enhance the GLD mechanism as illustrated in \Cref{fig:GLE}b. Given a raw feature $\mathbf{r} \in \mathbb{R}^{H \times W\times C}$, we first ``patchify'' $\mathbf{r}$ into $HW$ of $C$-dimensional patches, denoting as $\mathbf{r}' \in \mathbb{R}^{HW \times C}$. Each patch in $\mathbf{r}'$ can be considered a condensed vector for a certain receptive field in the original image. Our goal is to capture the correlations among these patches to facilitate the generation of GLDs. To achieve this, we adopt a standard transformer module to learn the relations among these vectors. A standard learnable positional encoding~\cite{ViT} is applied to $\mathbf{r}'$ before the transformer module.

After the transformer module, we reshape the feature back to $H \times W \times C$. Then a point-wise convolutional layer reduces the channel dimension to $K$, resulting in $\mathbf{q'} \in \mathbb{R}^{H\times W \times K}$. Remind that $K$ stands for the number of GLDs. Finally, a bottleneck module consisting of two linear layers and an activation function refines $\mathbf{q'}$ and predicts GLDs $\mathbf{q}$. In the implementation, we choose the Sigmoid function as the activation function, which maps $\mathbf{q}' \in [-\infty,\infty]$ into $\mathbf{q} \in [0,1]$.

Compared to the previous design, the enhanced GLE operates on the $C$-dimensional patch features rather than a single saliency map.
The patch features carry much more abundant local information and, therefore, enable the transformer module to better explore correlations among visual features at different locations.

\subsection{Layout Simulation and Semantic Augmentation}

In the preliminary paper, we extensively devised two types of augmentations, namely Layout simulation and Semantic augmentation (LS), with the aim of enhancing the extracted layout descriptors' quality and promoting the generalization of cross-view geo-localization models where:

\begin{figure*}[!t]
  \begin{center}
    \includegraphics[page=7,width=\textwidth, clip,trim=0cm 67cm 0cm 5cm]{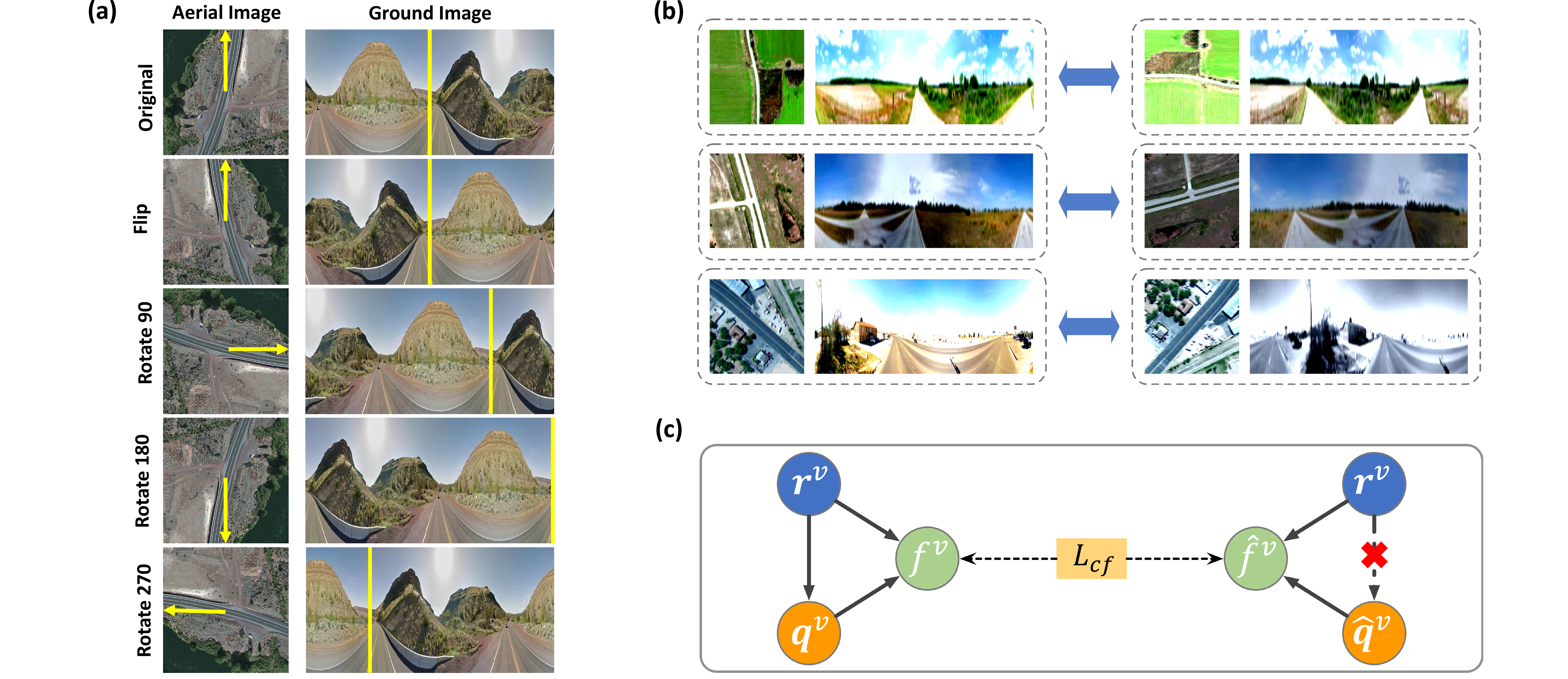}
  \end{center}
  \caption{(a) Illustration of the layout simulation. The left column is the aerial images and the right column is the ground images. The yellow arrows and lines indicate the north direction. (b) Three randomly sampled contrastive pairs from the CVUSA dataset. (c) Illustration of the proposed counterfactual learning schema. The arrows indicate the causal relation between two variables. The predicted feature $f^v$ and imaginary feature $\hat{f}^v$ are pushed away from each other by $L_{cf}$ (the dashed arrow) to provide weak supervision on raw feature $r^v$ and geometric layout descriptors $q^v$ to capture more distinctive geometric clues.}
  \label{fig:layout_sim}
\end{figure*}

\subsubsection{Layout Simulation}
It is a combination of a random flip and a random rotation ($90^{\circ}, 180^{\circ}, \text{ or } 270^{\circ}$)
that \textit{synchronously} applies to ground truth aerial and ground images. In this manner, low-level details are maintained, but the geometric layout is modified. As illustrated in~\Cref{fig:layout_sim}a, layout simulation can produce matched aerial-ground pairs with a different geometric layout.

\subsubsection{Semantic Augmentation}
randomly modifies the low-level features in aerial and ground images \textit{separately}. A color jitter is employed to modify the brightness, contrast, and saturation in images. Moreover,  random Gaussian blur, transform images to grayscale or posterized have been randomly applied. 

Unlike previous data augmentation methods, our LS does not break the geometric correspondence among visual features in the two views.
In this sense, LS generates new pairs with novel layouts from the original ones.
Let's denote the $i$th original pair of ground-aerial images as $P^{(i)}_{o}:=(I^g_i, I^a_i)$.
Formally, LS techniques can be expressed as a function that varies the 
input pair. Namely,
\begin{equation}
    P^{(i)}_{\gamma} = \phi(P^{(i)}_{o}; \theta_{\gamma}),
    \label{eq:LS}
\end{equation}
where $\theta_{\gamma}$ is the parameter for generating the LS-augmented pair $P^{(i)}_{\gamma}$, for example, rotating angles, horizontal flipping, and color jitter values.

\subsection{Contrastive Hard Samples Generation}
\label{sec::chsg}
In the current work, we have a closer look at the LS-augmented training dataset $\mathcal{T}_{LS}$. Specifically, we identify a subset for each raw aerial-ground pair $P^{(i)}_{o}$ which is defined as $S_{i}=\{P^{(i)}_{\gamma},P^{(i)}_{\delta},\dots\}$. Each element in $S_i$ is an LS-augmented pair from $P^{(i)}_{o}$. It is easy to see that the LS-augmented training dataset, $\mathcal{T}_{LS}$, is a union of such subsets. In practice, we intentionally choose $\theta_\gamma$ in \Cref{eq:LS} so that the generated contrastive pairs share different geometric layouts.
Subset $S_i$ is of particular interest as its elements are mutually hard samples in the sense that their visual features 
correspond to the same original visual features.

In our preliminary paper, 
each pair in a training batch is sampled from a distinct $S_{i}$. In this manner, the contrast between elements in a $S_i$ only happens at the inter-batch level, resulting inefficient exploitation of those hard samples. To tackle such inefficiency, we propose Contrastive Hard Samples Generation (\chsg{}). Remind that each element in $S_i$ is a hard sample for another. By leveraging this property, \chsg{} constructs a training batch as $\{P^{(1)}_{\gamma}, \dots,P^{(bs)}_{\gamma}, P^{(1)}_{\delta}, \dots, P^{(bs)}_{\delta}\}$, where $bs$ is the batch size and $P^{(i)}_{\gamma},P^{(i)}_{\delta}$ are both sampled from the $i$-th $S_{i}$.
Through this sample generation scheme, the intra-batch contrast among hard samples is emphasized. \Cref{fig:layout_sim}b visualizes three randomly sampled contrastive aerial-ground pairs from the CVUSA dataset.

\subsection{Counterfactual-based Learning Schema}
\label{sec::cf}
Due to the absence of ground truth geometric layout descriptors, the sub-module GLE would only receive indirect and insufficient supervision during training. 
Inspired by~\cite{fine_grained_cf}, we propose a counterfactual-based (CF-based) learning process. Specifically, we apply an intervention $do(\mathbf{q}^v=\hat{\mathbf{q}}^v)$ which substitutes $\mathbf{q}^v$ for a set of imaginary layout descriptors $\hat{\mathbf{q}}^v$ in~\Cref{eq:main}. This results in an imaginary representation $\hat{f}^v$.
Elements of $\hat{\mathbf{q}}^v$ are drawn from the uniform distribution $U[0,1]$. This process is illustrated in~\Cref{fig:layout_sim}c. In order to penalize $\hat{\mathbf{q}}^v$ and encourage $\mathbf{q}^v$ to capture more distinctive geometric clues, we maximize the distance between $f^v$ and $\hat{f}^v$ by minimizing our proposed counterfactual loss
\begin{equation}
\label{eq:cf}
    L^v_{cf} = \log\left(1+e^{-\beta^v \left[ d\left(f^v,\hat{f}^v\right) \right]}\right),
\end{equation}
where $\beta^v$ is a parameter to tune the convergence rate. The counterfactual loss provides a weakly supervision signal to the layout descriptors $\mathbf{q}$ via penalizing the imaginary descriptors $\hat{\mathbf{q}}$. In this way, the model can be away from apparently ``wrong'' solutions and learn a better latent feature representation.

\subsection{Training Objectives}
Besides the counterfactual loss, we also adopt the weighted soft margin triplet loss which pushes the matched pairs closer and unmatched pairs further away from each other
\begin{equation}
\label{eq:smtl}
    L_{triplet} = \log{\left( 1 + e^{\alpha\left[d(f^g_m, f^a_m)-d(f^g_m, f^a_n)\right]} \right)},
\end{equation}
where $\alpha$ is a hyperparameter that controls the convergence of training. $m,n \in \{1,2, \dots, N\}$ and $m \neq n$. Our final loss is
\begin{equation}
    L=L_{triplet}+L^a_{cf}+L^g_{cf}.
    \label{eq:loss}
\end{equation}

%% file: text/experiments.tex
\section{experiments}
\label{sec::experiment}

\subsection{Experiment Settings
\label{sec:expt_setting}}

\begin{figure}
    \centering
    \includegraphics[page=10,width=0.48\textwidth, clip,trim=3cm 30cm 10cm 42cm]{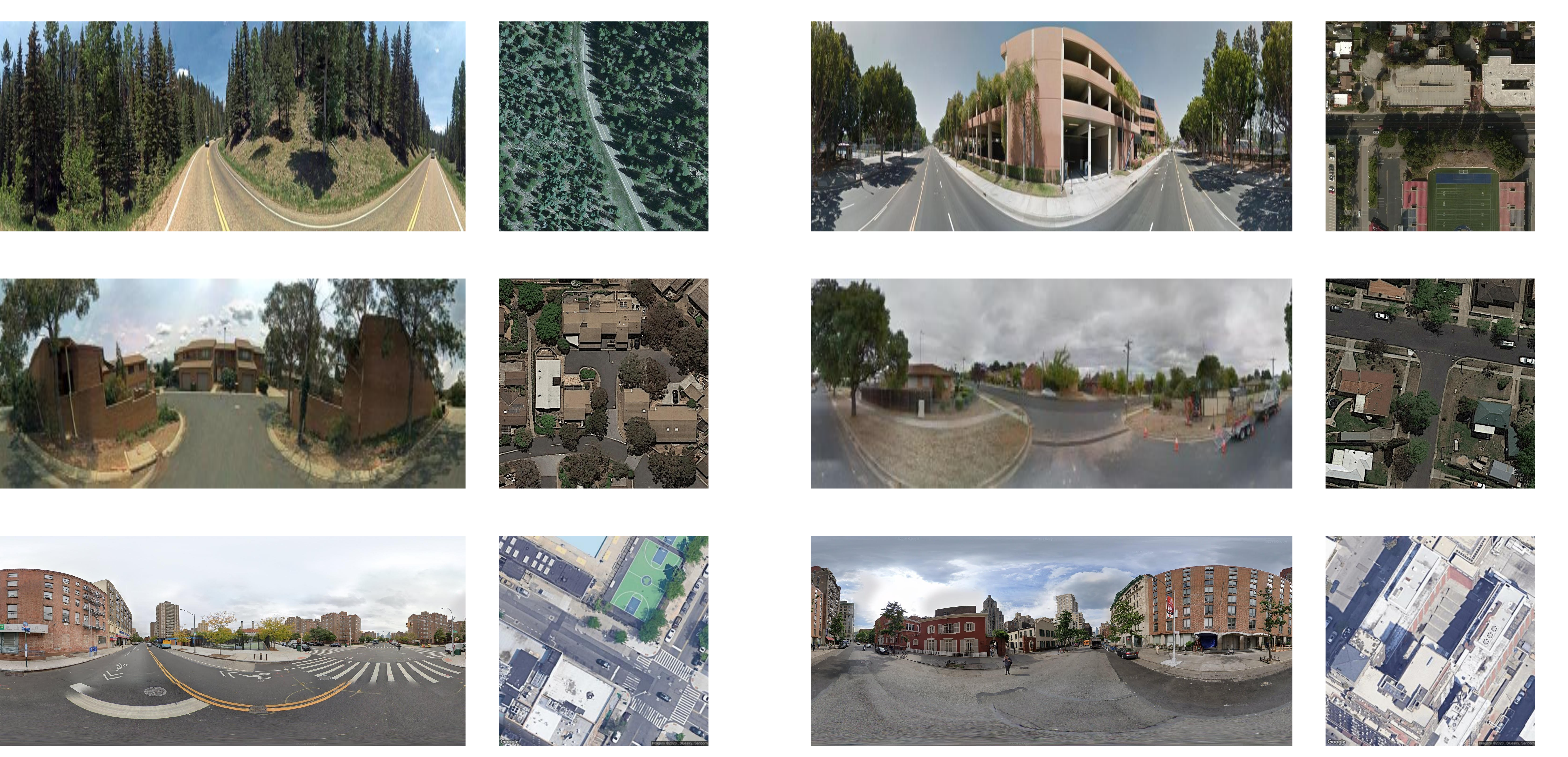}
    \caption{Six sample aerial-ground pairs from our training data. The top two pairs are from CVUSA~\cite{CVUSA} dataset. The middle two pairs are from CVACT~\cite{liu2019lending} dataset. The bottom two pairs are from VIGOR~\cite{Vigor} dataset.}
    \label{fig:dataset_samples}
\end{figure}

\begin{figure}[!t]
    \centering\includegraphics[width=0.48\textwidth]{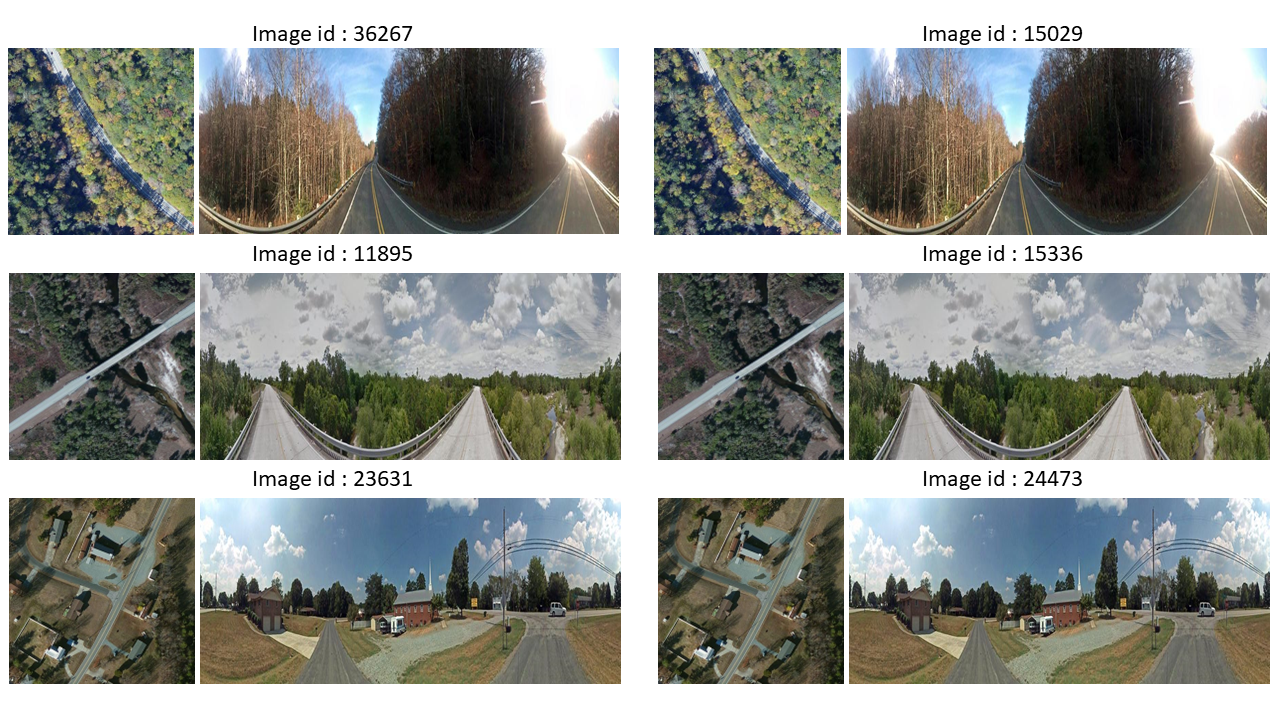}
    \caption{Three examples of repeated ground-aerial pairs from CVUSA training set.}
    \label{fig:duplicate}
\end{figure}

\paragraph{Dataset}
To evaluate the effectiveness of \ourmodel{}, we conduct extensive experiments on three datasets, CVUSA~\cite{CVUSA}, CVACT~\cite{liu2019lending}, and VIGOR~\cite{Vigor}.
Both CVUSA and CVACT contain $35,532$ training pairs. CVUSA provides $8,884$ pairs for testing and CVACT has the same number of pairs in its validation set (CVACT\_val). Besides, CVACT provides a challenging and large-scale testing set (CVUSA\_test) which contains $92,802$ pairs. VIGOR is a recently proposed challenging CVGL dataset. Different from CVUSA and CVACT, VIGOR does not assume the one-to-one match and center alignment in aerial-ground pairs. It collects $90,618$ ground panorama images and $105,124$ from 4 major cities in the U.S.A, including, New York, Chicago, Seattle, and San Francisco. Specifically, VIGOR elaborates two protocols, same-area (training and testing on 4 cities) and cross-area (training on New York and Seattle, testing on San Francisco and Chicago) to evaluate the comprehensive performance of CVGL models in metropolitan areas. We sample six aerial-ground image pairs from these three datasets and visualize them in~\Cref{fig:dataset_samples}. To be noticed that we identify $762$ and $43$ repeated pairs in the original training set and testing set of the CVUSA~\cite{CVUSA} dataset, respectively. We apply the md5 hash algorithm on the pixel values of each image to identify these repeated pairs. Three random examples with image IDs are presented in \Cref{fig:duplicate}. For fair comparisons with other baseline methods in this section, we remove the repeated pairs in the training set but \textit{keep} those in the testing set.

\paragraph{Evaluation Metric}
Similar to existing methods~\cite{SAFA,comingDown,l2ltr,cvmnet,liu2019lending,DSM}, we choose to use recall accuracy at top $K$ (R@$K$) for evaluation purposes. R@$K$ measures the probability of the ground truth aerial image ranking within the first $K$ predictions given a query image. In the following experiments, we evaluate the performance of all methods on \Rone{}, \Rfive{}, \Rten{}, and \Ronep{}.
\paragraph{Implementation Detail} We set $\alpha$ and $\beta$ in \Cref{eq:loss} to $10$ and $5$ respectively. The number of transformer encoder layers in the proposed GLE is set to $2$ and each with $4$ heads. The number of descriptors $K$ in \Cref{eq::f_product} is $8$. The model is trained on a single Nvidia V100 GPU for $200$ epochs with AdamW~\cite{adamw} optimizer and a learning rate of $10^{-4}$. The ground images and aerial images are resized to $122 \times 671$ and $256 \times 256$, respectively. Similar to previous methods~\cite{SAFA,DSM,l2ltr,comingDown,cvmnet}, we set the batch size to $32$. Within each batch, the exhaustive mini-batch strategy~\cite{Vo} is utilized for constructing triplet pairs. Considering the recently published CVGL models adopt a more advanced backbone (i.e. L2LTR~\cite{l2ltr} uses ViT~\cite{ViT} and TransGeo leverages DeiT~\cite{Deit}), we upgrade the backbone from our preliminary version of ResNet-34~\cite{resnet} to ConvNeXt-T~\cite{convnext} which is more advanced but has a similar number of trainable parameters. Notice that ConvNeXt-T is a lightweight model and does not outperform ViT~\cite{ViT} and DeiT~\cite{Deit} on ImageNet~\cite{imagenet} benchmark.
We take the output before the last average pooling layer of ConvNeXt-T as the raw feature $\mathbf{r}^v$ in \Cref{eq:main}, the latent feature $f^v$ is a $8 \times 384 = 3072$ dimensional vector, where $8$ is $K$ and $384$ is the number of channels in $r^v$. All the experiments are performed under the settings mentioned above unless we specify otherwise.

\subsection{Main results}
\label{sec::main_results}

\input{tables/same_area}

\subsubsection{CVUSA and CVACT same-area experiment}
The same-area evaluation results of the proposed \ourmodel{} and current State-Of-The-Art (SOTA) methods on CVUSA~\cite{CVUSA} and CVACT~\cite{liu2019lending} are presented in \Cref{tab:same_area}. The best result is indicated in the magenta text, while the second-best result is indicated in the cyan text. As illustrated in \Cref{tab:same_area}, the proposed \ourmodel{} achieves SOTA accuracy on CVACT benchmarks while using the polar transformation. Specifically, on the CVACT\_val, \ourmodel{} improves from $86.21\%$ to $87.61\%$ on \Rone{}. On one of the most challenging benchmarks, CVACT\_test, the proposed model improves from $64.52\%$ to $67.57\%$ on \Rone{}. On the CVUSA benchmark, the accuracy of \ourmodel{} is very close to the SOTA value. Without polar transformation, \ourmodel{} achieves SOTA performance on \Rone{} of all three benchmarks. \ro{Notably, on CVACT\_val and CVACT\_test, our \ourmodel{} without polar transformation not only achieves SOTA performance but also outperforms other models trained with polar transformation, which is not observable in other methods such as SAFA~\cite{SAFA} and L2LTR~\cite{l2ltr}. \ro{We attribute this to} the ability of the proposed novel Geometric Layout Extractor (GLE) to explore global spatial configurations.}

\input{tables/cross_area}

\subsubsection{CVUSA and CVACT cross-area experiment}
As we discussed in \Cref{sec::introduction}, the generalization of the CVGL models is an important property. Therefore, we conduct a cross-area experiment as presented in \Cref{tab:corss_area}. This experiment includes two tasks, training on CVUSA and then testing on CVACT (CVUSA $\rightarrow$ CVACT) and training on CVACT and then testing on CVUSA (CVACT $\rightarrow$ CVUSA). Firstly, as indicated in \Cref{tab:corss_area}, \ourmodel{} achieves SOTA performance on all evaluation metrics, both when trained with polar transformation and without polar transformation. Specifically, on CVUSA $\rightarrow$ CVACT task, \ourmodel{} achieves $61.17\%$ and $60.16\%$ \Rone{} accuracy with or without polar transformation which are a significant improvement over the previous SOTA ($8\%$ and $16\%$ respectively). More importantly, we improve \Rone{} in CVACT $\rightarrow$ CVUSA task from previous SOTA $44.07\%$ to $53.89\%$ and from $29.85\%$ to $52.56\%$ on trained with polar transformation and without polar transformation respectively.

\input{tables/vigor}

\subsubsection{VIGOR experiment}
To further investigate the performance of \ourmodel{} under more challenging and realistic settings, we benchmark the proposed model on the VIGOR~\cite{Vigor} dataset. Different from CVUSA~\cite{CVUSA} and CVACT~\cite{liu2019lending} datasets which maintain a one-to-one correspondence between aerial image and ground image, each ground query image in VIGOR~\cite{Vigor} has two or more corresponding aerial images. This setting makes VIGOR~\cite{Vigor} closer to real-world scenarios. Due to the many-to-one correspondence, it is considered one of the most challenging datasets in CVGL.
\Cref{tab:vigor} presents the performance comparison between the proposed \ourmodel{} and other baseline methods. \ourmodel{} achieves the SOTA performance by improving from the previous SOTA $22.35\%$ to $36.01\%$ which is a $13.66\%$ increasing in cross-area experiment. Although, \ourmodel{} does not achieve the best result on same-area evaluation. However, the performance gap in the same-area experiment is relatively small considering the improvement in the cross-area experiment. \common{The same-area performance difference might come from both model scale and the augmentation in \chsg{} such as the color space distortion and geometric transformations. In summary, \ourmodel{} shows significantly enhances the cross-area performance over the SOTA methods on cross-area evaluation. Although, there is a slight drop in same-area evaluation. However, in later~\Cref{sec::computational_efficiency}, we show that \ourmodel{} is the smallest model among all the baseline methods. Thus 
the same-area performance can be further improved by applying a larger backbone. However, it is not the focus of this paper.} In the next section, we conduct ablation studies to demonstrate the effectiveness of the number of descriptors, the proposed GLE, and \chsg{} on our model.

\subsection{Ablation studies}

In this subsection, we explore the effectiveness of each component in \ourmodel{}. It is worth mentioning that the contribution of CF learning schema had been studied in detail in our previous work~\cite{GeoDTR}. Therefore, we leave the discussion about CF to the Appendix~(\Cref{sec::ablation_CF}).

\input{tables/GLE}
\subsubsection{Ablation study of GLE, \chsg{}, and Backbone}
To investigate the attribute of each component in the \ourmodel{}, we conduct an ablation study of the proposed GLE (\Cref{sec::GLE}) and \chsg{} (\Cref{sec::chsg}). The results are presented in \Cref{tab:GLE}. All experiments in \Cref{tab:GLE} are trained with LS techniques and polar transformation by default. We vary the combination of the GLE, \chsg{}, and backbones. As shown in \Cref{tab:GLE}, we first can observe the difference by comparing the proposed GLE in this work (``v2'') and GLE from our preliminary work~\cite{GeoDTR} (``v1''). \ro{As we can see training with the novel GLE (v2) on ResNet-34~\cite{resnet} generally improves the cross-area performance on both CVUSA and CVACT datasets (except while training with \chsg{} on the CVACT, there is a slight drop from $50.63\%$ to $49.09\%$). For example, the \Rone{} accuracy improves from $57.03\%$ (line 1) to $59.72\%$ (line 3) on the CVUSA dataset.} By comparing each pair of experiments training with and without \chsg{}, we observe that \chsg{} significantly enhances the performance on cross-area benchmarks while having minimal impact on the same-area performance. For instance, while training on ConvNeXt-T with the proposed GLE, training without \chsg{} only achieves $42.11\%$ on the CVACT dataset. While training with \chsg{}, this number increases to $53.89\%$. Lastly, we find that upgrading the backbone slightly improves the same-area performance. It is noticeable that the cross-area performance degrades if training with the new GLE and without \chsg{} (i.e. line 3 and line 5 in both CVUSA and CVACT experiment from \Cref{tab:GLE}). Once training with \chsg{}, this degradation can be mitigated due to the strong supervision signal from the additional hard aerial-ground pairs. To reveal the underlying mechanism of the proposed GLE and \chsg{}, We provide more complete investigations as well as more detailed experiments and analyses in~\Cref{sec::discussion}. \common{For more ablation studies on VIGOR dataset, please refer to our Appendix(\Cref{sec:ablation_vigor}).}

\input{tables/descriptors}
\subsubsection{Varying the Number of Geometric Layout Descriptors}
CVGL is sensitive to the latent feature dimensions in a real-world deployment. Smaller latent feature dimensions shorten the inference time and require less reference feature storage space. In our proposed \ourmodel{}, we can change the latent dimensions by varying the number of GLDs. The result is illustrated in \Cref{tab:descriptors}. It is clear to see that with the smaller latent feature dimensions, both the same-area and cross-area performance degrade. However, to be noticed, even with $2$ descriptors and $768$ latent dimensions, our model still achieves SOTA performance on cross-area benchmarks. More importantly, the degradation of the same-area result does not exceed $2\%$ which means the model is still comparable to the existing SOTA result on same-area benchmarks. For more ablation studies of the Geometric Layout Descriptors, please refer to our Appendix~(\Cref{sec::ablation_GLD}).

\input{tables/CHSG-1}
\subsubsection{Effect of CHSG over LS}
To better evaluate the effect of \chsg{} on the CVGL performance, in addition to the plain LS, we conduct experiments with LS, \chsg{}, and various batch sizes, as shown in~\Cref{tab:chsg_LS_batch}. We first note that different configurations share similar same-area performance. However, without LS techniques and \chsg{}, the model performs poorly on cross-area benchmarks. We also observe that a larger batch size ($64$) with LS techniques can have a marginal performance improvement in the same-area evaluation compared with the standard batch size ($32$) with LS techniques. Still, the cross-area performance remains at a similar level compared with the standard batch size. More importantly, compared with other configurations, training with \chsg{} can significantly boost the cross-area performance, achieving $61.17\%$ on \Rone{}, while maintaining a batch size of $32$. This demonstrates that simply increasing the batch size can hardly improve both same-area and cross-area performance. But \chsg{} can efficiently guide the model to learn more distinct geometric patterns which is helpful in improving cross-area performance. 

\Cref{sec::discussion} includes additional fine-grained experiments to further demonstrate 
the contributions in \chsg{} at inter- and intra-batch levels. For the performance of our \chsg{} on the recently proposed SAIG~\cite{SAIG} model, please refer to our Appendix(\Cref{sec:arch_compare_saig}). 

\input{tables/computational_efficiency}
\subsection{Computational Efficiency}
\label{sec::computational_efficiency}
To better understand the proposed \ourmodel{} in inference efficiency, we compare the number (\#) of the trainable parameters, inference time, pretrained model, and computational cost for \ourmodel{} and recently published methods as presented in \Cref{tab:computation}. It is clear to see that \ourmodel{} has the least \# of trainable parameters because of the new GLE design and the ConvNeXt-T~\cite{convnext} backbone. Compared with GeoDTR~\cite{GeoDTR} and TransGeo~\cite{transgeo}, \ourmodel{} has $50\%$ and $45\%$ less trainable parameters respectively. Our model has similar inference time and floating point operations per second to TransGeo~\cite{transgeo} which is one of the lightest models in CVGL. Compared with L2LTR~\cite{l2ltr}, \ourmodel{} does not require a large pretrained backbone such as ViT which is pretrained on ImageNet-21K~\cite{imagenet}.

%% file: tables/same_area.tex
\begin{table*}[!ht]
    \centering
    \setlength{\tabcolsep}{1.5 mm}
    \caption{\label{tab:same_area}Comparison between our \ourmodel{} and baseline methods on CVUSA, CVACT\_val, and CVACT\_test benchmarks. \B{Magenta} text stands for the best results and \SB{cyan} text stands for the second best result.}
    \begin{tabular}{ccccccccccccc}
    \toprule 
    \multirow{2}{*}{Method} & \multicolumn{4}{c}{CVUSA} & \multicolumn{4}{c}{CVACT\_val} &
    \multicolumn{4}{c}{CVACT\_test} \\
    \cmidrule(lr){2-5}
    \cmidrule(lr){6-9}
    \cmidrule(lr){10-13}
     & \Rone{} & \Rfive{} & \Rten{} & \Ronep & \Rone{} & \Rfive{} & \Rten{} & \Ronep & \Rone{} & \Rfive{} & \Rten{} & \Ronep \\
     \multicolumn{13}{l}{\textit{w/ PT}} \\ \midrule
    CVFT~\cite{featureTransport} & 61.43\% & 84.69\% & 90.49\% &  99.02\% &  61.05\% & 81.33\% & 86.52\% & 95.93\% & 26.12\% & 45.33\% & 53.80\% & 71.69\% \\
    SAFA~\cite{SAFA} & 89.84\% & 96.93\% & 98.14\% & 99.64\% & 81.03\% & 92.80\% & 94.84\% & 98.17\% & 55.50\% & 79.94\% & 85.08\% & 94.49\% \\
    DSM~\cite{DSM} & 91.93\% & 97.50\% & 98.54\% & 99.67\% & 82.49\% & 92.44\% & 93.99\% & 97.32\% & 35.63\% & 60.07\% & 69.10\% & 84.75\% \\
    CDE~\cite{comingDown} & 92.56\% & 97.55\% & 98.33\% & 99.57\% & 83.28\% & 93.57\% & 95.42\% & 98.22\% & 61.29\% & 85.13\% & 89.14\% & 98.32\% \\
    L2LTR~\cite{l2ltr} & 94.05\% & 98.27\% & 98.99\% & 99.67\% & 84.89\% & 94.59\% & 95.96\% & 98.37\% & 60.72\% & 85.85\% & 89.88\% & 96.12\% \\
    \rt{TGCNN~\cite{TGCNN}} & 94.15\% & 98.21\% & 98.94\% & \SB{99.79\%} & 84.92\% & 94.46\% & 95.88\% & 98.36\% & - & - & - & - \\
    \rt{MGTL~\cite{MGTL}} & 94.50\% & 98.41\% & \SB{99.20\%} & 99.78\% & 85.42\% & 94.64\% & 96.11\% & \SB{98.51\%} & 61.55\% & 86.61\% & 90.74\% & 98.46\% \\
    SEH~\cite{SEH} & 95.11\% & 98.45\% & 99.00\% & 99.78\% & 84.75\% & 93.97\% & 95.46\% & 98.11\% & - & - & - & - \\
    \ro{SAIG~\cite{SAIG}} & 92.71\% & 97.92\% & 98.89\% & 99.71\% & 84.42\% & 94.09\% & 95.57\% & 98.49\% & - & - & - & - \\
    GeoDTR~\cite{GeoDTR} & \B{95.43\%} & \B{98.86\%} & \B{99.34\%} & \B{99.86\%} & \SB{86.21\%} & \SB{95.44\%} & \B{96.72\%} & \B{98.77\%} & \SB{64.52\%} & \SB{88.59\%} & \SB{91.96\%} & \SB{98.74\%} \\
    \ourmodel{} (ours) & \SB{95.40\%} & \SB{98.44\%} & 99.05\% & 99.75\% & \B{87.61\%} & \B{95.48\%}	& \SB{96.52\%} & 98.34\% & \B{67.57\%} & \B{89.84\%} & \B{92.57\%} & \B{98.54\%} \\[5pt]
    \multicolumn{13}{l}{\textit{w/o PT}} \\ \midrule
    CVM-Net~\cite{cvmnet} & 22.47\% & 49.98\% &  63.18\% &  93.62\% &  20.15\% &  45.00\% &  56.87\% &  87.57\% &  5.41\% &  14.79\% &  25.63\% &  54.53\% \\
    Liu \& Li & 40.79\% &   66.82\% &   76.36\% &   96.12\% &   46.96\% &   68.28\% &   75.48\% &   92.01\% &    19.21\% &   35.97\% &   43.30\% &   60.69\% \\
    SAFA~\cite{SAFA} & 81.15\% & 94.23\% & 96.85\% & 99.49\% & 78.28\% & 91.60\% & 93.79\% & 98.15\% & - & - & - & - \\
    L2LTR~\cite{l2ltr} & 91.99\% & 97.68\% & 98.65\% & 99.75\% & 83.14\% & 93.84\% & 95.51\% & \SB{98.40\%} & 58.33\% & 84.23\% & 88.60\% & 95.83\% \\
    TransGeo~\cite{transgeo} & \SB{94.08\%} & 98.36\% &  \SB{99.04\%} & \SB{99.77\%} & 84.95\% & 94.14\% &  95.78\% & 98.37\% & - & - & - & - \\
    \ro{SAIG~\cite{SAIG}} & 92.71\% & 97.92\% & 98.89\% & 99.71\% & 84.42\% & 94.09\% & 95.57\% & \B{98.49\%} & - & - & - & - \\
    GeoDTR~\cite{GeoDTR} & 93.76\% & \B{98.47\%} & \B{99.22\%} & \B{99.85\%} & \SB{85.43\%} & \SB{94.81\%} & \SB{96.11\%} & 98.26\% & \SB{62.96\%} & \SB{87.35\%} & \SB{90.70\%} & \SB{98.61\%} \\
    \ourmodel{} (ours) & \B{95.05\%} & \SB{98.42\%} & 98.92\% & \SB{99.77\%} & \B{87.76\%} & \B{95.50\%}	& \B{96.50\%} & 98.32\% & \B{67.75\%} & \B{90.15\%} & \B{92.73\%} & \B{98.53\%} \\
    \bottomrule
    \end{tabular}
\end{table*}

%% file: tables/cross_area.tex
\begin{table*}[!ht]
    \centering
    \caption{\label{tab:corss_area} Cross-area comparison between the proposed \ourmodel{} and baselines on CVUSA and CVACT benchmarks. \B{Magenta} text stands for the best results and \SB{cyan} text stands for the second best result.}
    \begin{tabular}{ccccccccc}
    \toprule 
    \multirow{2}{*}{Method/Task} & \multicolumn{4}{c}{CVUSA $\rightarrow$ CVACT} & \multicolumn{4}{c}{CVACT $\rightarrow$ CVUSA} \\
    \cmidrule(lr){2-5}
    \cmidrule(lr){6-9}
     & \Rone{} & \Rfive{} & \Rten{} & \Ronep & \Rone{} & \Rfive{} & \Rten{} & \Ronep \\
    \multicolumn{9}{l}{\textit{w/ PT}} \\ \midrule
    SAFA~\cite{SAFA} & 30.40\% & 52.93\% & 62.29\% & 85.82\% & 21.45\% & 36.55\% & 43.79\% & 69.83\% \\
    DSM~\cite{DSM} &33.66\% & 52.17\% & 59.74\% & 79.67\% & 18.47\% & 34.46\% & 42.28\% & 69.01\% \\
    L2LTR~\cite{l2ltr} & 47.55\% & 70.58\% & 77.39\% & 91.39\% & 33.00\% & 51.87\% & 60.63\% & 84.79\% \\
    GeoDTR~\cite{GeoDTR} & \SB{53.16\%} & \SB{75.62\%} & \SB{81.90\%} & \SB{93.80\%} & \SB{44.07\%} & \SB{64.66\%} & \SB{72.08\%} & \SB{90.09\%} \\
    \ourmodel{} (ours) & \B{61.17\%} & \B{80.22\%} & \B{85.45\%} & \B{94.56\%} & \B{53.89\%} & \B{74.56\%} & \B{81.10\%} & \B{94.93\%} \\ [5pt]
    \multicolumn{9}{l}{\textit{w/o PT}} \\ \midrule
    TransGeo~\cite{transgeo} & 37.81\% & 61.57\% & 69.86\% & 89.14\% & 17.45\% & 32.49\% & 40.48\% & 69.14\% \\
    \ro{SAIG~\cite{SAIG}} & 15.29\% & 33.07\% & 42.14\% & 72.95\% & 18.97\% & 35.60\% & 44.28\% & 75.33\% \\
    GeoDTR~\cite{GeoDTR} & \SB{43.72\%} & \SB{66.99\%} & \SB{74.61\%} & \SB{91.83\%} & \SB{29.85\%} & \SB{49.25\%} & \SB{57.11\%} & \SB{82.47\%} \\
    \ourmodel{} (ours) & \B{60.16\%} & \B{79.97\%} & \B{84.67\%} & \B{94.48\%} & \B{52.56\%} & \B{73.08\%} & \B{79.82\%} & \B{94.80\%} \\
    \bottomrule
    \end{tabular}
\end{table*}

%% file: tables/vigor.tex
\begin{table*}[!ht]
    \centering
    \caption{\label{tab:vigor} Comparison between the proposed \ourmodel{} and baseline methods on VIGOR benchmark. \B{Magenta} text stands for the best results and \SB{cyan} text stands for the second best result.}
    \begin{tabular}{ccccccccccc}
    \toprule 
    & \multicolumn{5}{c}{Same-area} & \multicolumn{5}{c}{Cross-area} \\
    \cmidrule(lr){2-6}
    \cmidrule(lr){7-11}
     Method & \Rone{} & \Rfive{} & \Rten{} & \Ronep & Hit Rate & \Rone{} & \Rfive{} & \Rten{} & \Ronep & Hit Rate \\
    \midrule
    SAFA~\cite{SAFA} & 18.69\% & 43.64\% & 55.36\% & 97.55\% & 21.90\% & 2.77\% & 8.61\% & 12.94\% & 62.64\% & 3.16\% \\
    SAFA+Mining~\cite{Vigor} & 38.02\% & 62.87\% & 71.12\% & 97.63\% & 41.81\% & 9.23\% & 21.12\% & 28.02\% & 77.84\% & 9.92\% \\
    VIGOR~\cite{Vigor} & 41.07\% & 65.81\% & 74.05\% & 98.37\% & 44.71\% & 11.00\% & 23.56\% & 30.76\% & 80.22\% & 11.64\% \\
    TransGeo~\cite{transgeo} & \B{61.48\%} & \B{87.54\%} & \B{91.88\%} & \B{99.56\%} & \B{73.09\%} & 18.99\% & 38.24\% & 46.91\% & 88.94\% & 21.21\% \\
    \ro{SAIG~\cite{SAIG}} & 55.60\% & 81.63\% & - & \SB{99.43\%} & 63.57\% & 22.35\% & 42.43\% & - & 90.83\% & 24.69\% \\
    GeoDTR~\cite{GeoDTR} & 56.51\% & 80.37\% & 86.21\% & 99.25\% & 61.76\% & \SB{30.02\%} & \SB{52.67\%} & \SB{61.45\%} & \SB{94.40\%} & \SB{30.19\%} \\
    \ourmodel{} (ours) & \SB{59.01\%} & \SB{81.77\%} & \SB{87.10\%} & 99.07\% & \SB{67.41\%} & \B{36.01\%} & \B{59.06\%} & \B{67.22\%} & \B{94.95\%} & \B{39.40\%} \\
    \bottomrule
    \end{tabular}
\end{table*}

%% file: tables/GLE.tex
\begin{table*}[!ht]
    \centering
    \caption{\label{tab:GLE} Ablation study of the proposed Geometric Layout Extractor (GLE) and \chsg{} on CVUSA and CVACT dataset with same-area and cross-area evaluation. In the GLE configuration, ``v1'' stands for the GLE from our preliminary work~\cite{GeoDTR}. ``v2'' stands for the GLE proposed in this work. ``Same-area'' is training and testing on the same benchmark. ``Cross-area'' is training and testing on different benchmarks (i.e. CVUSA $\rightarrow$ CVACT or CVACT $\rightarrow$ CVUSA).}
    \begin{tabular}{cc@{\hspace*{2mm}}c@{\hspace*{2mm}}ccccccccc}
    \toprule 
     \multirow{2}{*}{\ro{Training Set}} & \multicolumn{3}{c}{Configurations} & \multicolumn{4}{c}{Same-area} & \multicolumn{4}{c}{Cross-area} \\
    \cmidrule(lr){2-4}
    \cmidrule(lr){5-8}
    \cmidrule(lr){9-12}
     & \chsg{} & GLE & Backbone & \Rone{} & \Rfive{} & \Rten{} & \Ronep  & \Rone{} & \Rfive{} & \Rten{} & \Ronep \\ \midrule
    \multirow{6}{*}{\rotatebox[origin=c]{90}{CVUSA}} \
    & $\times$ & v1 & ResNet-34 & 95.43\% & 98.86\% & 99.34\% & 99.86\% & 53.16\% & 75.62\% & 81.90\% & 93.80\%  \\
    & $\checkmark$ & v1 & ResNet-34 & 94.97\% & 98.47\% & 98.98\% & 99.73\% & 57.03\% & 76.92\% & 82.77\% & 93.87\% \\
    & $\times$ & v2 & ResNet-34 & 94.88\% & 98.37\% & 98.99\% & 99.76\% & 54.05\% & 75.93\% & 82.58\% & 93.94\%\\
    & $\checkmark$ & v2 & ResNet-34 & 94.26\% & 98.21\% & 98.91\% & 99.71\% & 59.72\% & 78.78\% & 82.95\% & 92.37\% \\
    & $\times$ & v2 & ConvNeXt-T & 95.77\% & 98.93\% & 99.36\% & 99.81\% & 53.65\% & 74.91\% & 81.39\% & 93.27\%  \\
    & $\checkmark$ & v2 & ConvNeXt-T & 95.40\% & 98.44\% & 99.05\% & 99.75\% & 61.17\% & 80.22\% & 85.45\% & 94.56\% \\ \midrule
    \multirow{6}{*}{\rotatebox[origin=c]{90}{CVACT}} 
    & $\times$ & v1 & ResNet-34 & 86.21\% & 95.44\% & 96.72\% & 98.77\% & 44.07\% & 64.66\% & 72.08\% & 90.09\%  \\
    & $\checkmark$ & v1 & ResNet-34 & 86.48\% & 95.26\% & 96.58\% & 98.26\% & 50.63\% & 71.26\% & 78.50\% & 93.87\% \\
    & $\times$ & v2 & ResNet-34 & 85.86\% & 95.06\% & 96.30\% & 98.38\% & 45.44\% & 66.07\% & 72.35\% & 90.97\%  \\
    & $\checkmark$ & v2 & ResNet-34 & 86.19\% & 94.90\% & 96.29\% & 98.30\% & 49.09\% & 70.65\% & 79.09\% & 93.42\% \\
    & $\times$ & v2 & ConvNeXt-T & 88.18\% & 95.57\% & 96.68\% & 98.67\% & 42.11\% & 62.00\% & 70.58\% & 90.88\%  \\
    & $\checkmark$ & v2 & ConvNeXt-T & 87.61\% & 95.48\% & 96.52\% & 98.34\% & 53.89\% & 74.56\% & 81.10\% & 94.93\% \\
    \bottomrule
    \end{tabular}
\end{table*}

%% file: tables/descriptors.tex
\begin{table*}[!ht]
    \centering
    \caption{\label{tab:descriptors} Ablation study of different numbers of descriptors (2,4,6,8) and their corresponding latent feature dimensions on CVUSA and CVACT datasets. ``Same-area'' is training and testing on the same benchmark. ``Cross-area'' is training and testing on different benchmarks (i.e. CVUSA $\rightarrow$ CVACT or CVACT $\rightarrow$ CVUSA).}
    \begin{tabular}{ccccccccccc}
    \toprule 
    &  &  &
    \multicolumn{4}{c}{Same-area} & \multicolumn{4}{c}{Cross-area} \\
    \cmidrule(lr){4-7}
    \cmidrule(lr){8-11}
     \ro{Training Set}& \# of Des. & Dimension & \Rone{} & \Rfive{} & \Rten{} & \Ronep  & \Rone{} & \Rfive{} & \Rten{} & \Ronep \\\midrule
    \multirow{4}{*}{\rotatebox[origin=c]{90}{CVUSA}} & 
    8 & 3072 & 95.40\% & 98.44\% & 99.05\% & 99.75\% & 61.17\% & 80.22\% & 85.45\% & 94.56\% \\
    & 6 & 2304 & 95.09\% & 98.38\% & 99.04\% & 99.73\% & 60.73\% & 79.68\% & 84.97\% & 95.21\% \\
    & 4 & 1536 & 94.57\% & 98.32\% & 98.96\% & 99.67\% & 60.21\% & 79.43\% & 85.86\% & 94.27\% \\
    & 2 & 768 & 93.59\% & 98.08\% & 98.87\% & 99.65\% & 59.21\% & 78.96\% & 84.48\% & 94.06\% \\
    \midrule
    \multirow{4}{*}{\rotatebox[origin=c]{90}{CVACT}} &
    8 & 3072 & 87.61\% & 95.48\% & 96.52\% & 98.34\% & 53.89\% & 74.56\% & 81.10\% & 94.93\% \\
    & 6 & 2304 & 87.05\% & 95.33\% & 96.55\% & 98.41\% & 53.61\% & 73.97\% & 80.69\% & 94.57\% \\
    & 4 & 1536 & 86.85\% & 95.42\% & 96.62\% & 98.26\% & 52.75\% & 73.30\% & 80.36\% & 94.51\% \\
    & 2 & 768 & 86.22\% & 94.96\% & 96.37\% & 98.22\% & 50.68\% & 71.12\% & 78.89\% & 94.10\% \\
    \bottomrule
    \end{tabular}
\end{table*}

%% file: tables/CHSG-1.tex
\begin{table}[!ht]
    \centering
    \caption{\label{tab:chsg_LS_batch} Comparison of the proposed \ourmodel{} trained with different data sampling strategies with different batch sizes on the CVUSA dataset. ``RAW'' stands for the original aerial-ground pairs as input without any augmentation. ``BS'' is short for batch size. The model is trained on the CVUSA dataset and evaluated on both same-area and cross-area benchmarks.}
    \begin{tabular}{cc@{\hspace*{2mm}}ccccc}
    \toprule 
    & Configuration & BS & {\Rone{}} & {\Rfive{}} & {\Rten{}} & {\Ronep} \\\midrule
    \multirow{4}{*}{\rotatebox[origin=c]{90}{Same-area}}
    & Raw & 32 & 95.77\% & 98.93\% & 99.36\% & 99.81\% \\
    & LS & 32 & 95.90\% & 98.96\% & 99.42\% & 99.81\% \\
    & LS & 64 & 96.24\% & 99.04\% & 99.51\% & 99.84\% \\
    & CHSG & 32 & 95.40\% & 98.44\% & 99.05\% &	99.75\% \\\midrule
    \multirow{4}{*}{\rotatebox[origin=c]{90}{Cross-area}}
    & Raw & 32 & 53.65\% & 74.91\% & 81.39\% & 93.27\% \\
    & LS & 32 & 58.62\% & 78.53\% & 84.20\% & 94.09\% \\
    & LS & 64 & 57.18\% & 78.33\% & 84.14\% & 93.72\% \\
    & CHSG & 32 & 61.17\% & 80.22\% & 85.45\% & 94.56\% \\
    \bottomrule
    \end{tabular}
\end{table}

%% file: tables/computational_efficiency.tex
\begin{table*}[!t]
    \caption{\label{tab:computation} Computational efficiency comparison between the proposed \ourmodel{} and GeoDTR~\cite{GeoDTR}, TransGeo~\cite{transgeo}, L2LTR~\cite{l2ltr}. All experiments are conducted on a single Nvidia V100 GPU.}
    \centering
    \small
    \setlength{\tabcolsep}{3.5 mm}
    \begin{tabular}{ccccc}
    \toprule
     Method & \# of Parameters & Inference Time & Pretraining model & Computational Cost \\ \midrule
     L2LTR~\cite{l2ltr} & 195.9M & 405ms & ImageNet-21K for ViT & 46.77 GFLOPS \\
     TransGeo~\cite{transgeo} & 44.0M & 99ms & ImageNet-1K for DeiT & 11.35 GFLOPS \\
     \ro{SAIG~\cite{SAIG}} & 31.2M & 115ms & ImageNet-1K for SAIG & 13.30 GFLOPS \\
     GeoDTR~\cite{GeoDTR} & 48.5M & 235ms & ImageNet-1K for ResNet34 & 39.89 GFLOPS \\
     \ourmodel{} (ours) & 24.7M & 103ms & ImageNet-1K for ConvNeXt-Tiny & 11.25 GFLOPS \\
    \bottomrule
    \end{tabular}
\end{table*}

%% file: text/discussion.tex
\begin{figure*}
    \centering
    \includegraphics[page=9,width=\textwidth, clip,trim=0cm 72cm 18cm 0cm]{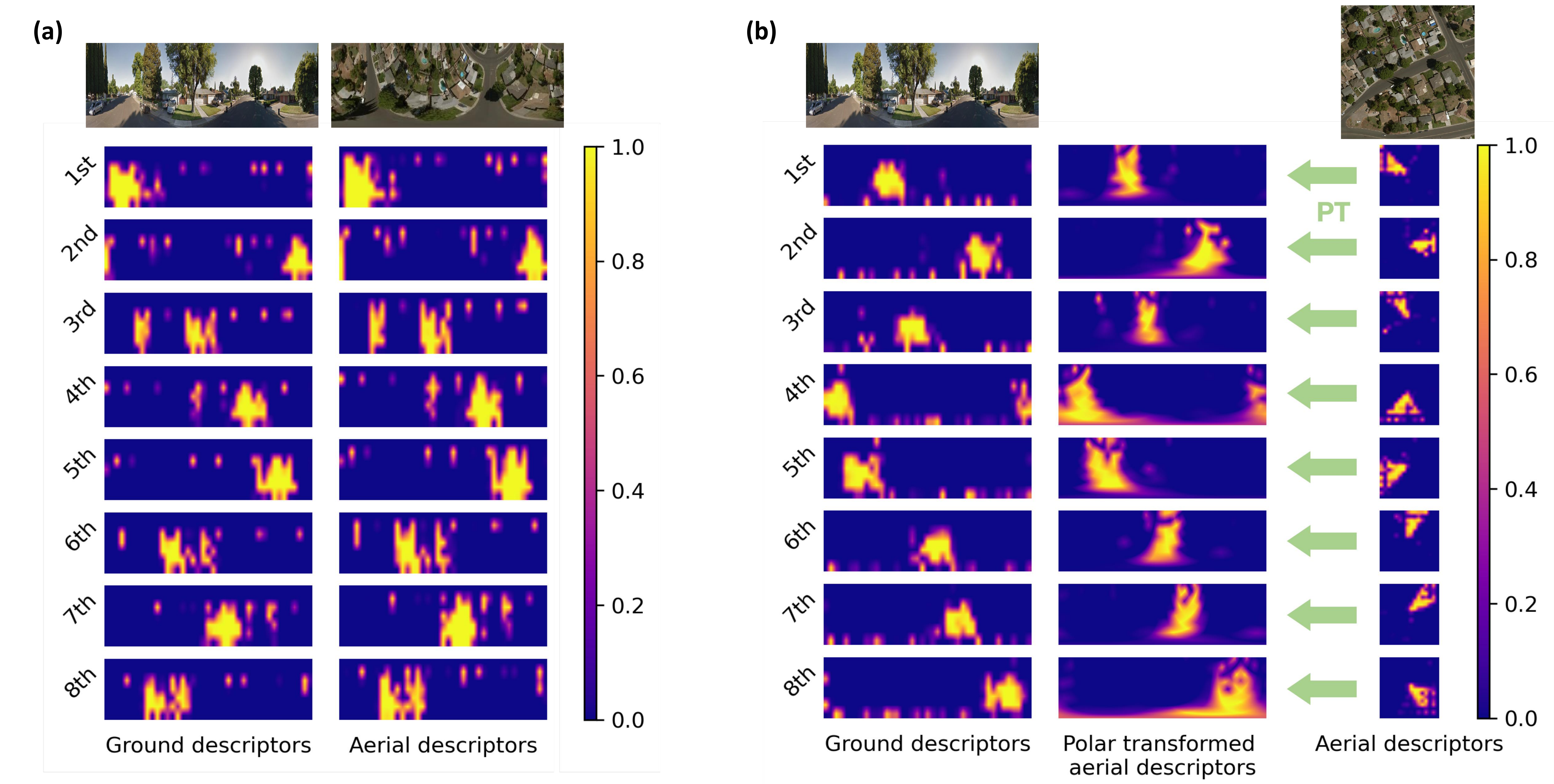}
    \caption{Visualization of the learned descriptors from the proposed \ourmodel{} training with polar transformation (a) and without polar transformation (b). The images at the top are the input ground images and aerial images accordingly. From top to bottom are the first to the eighth heatmap visualization of learned descriptors. For better visualization purposes, we apply the polar transformation on the aerial descriptors from the model training without polar transformation(b). We note a strong alignment of the learned descriptors in both (a) and (b) which demonstrates the ability of \ourmodel{} to capture the geometric correspondence. This also illustrates the reason that our model shows a similar performance while training with or without polar transformation.}
    \label{fig:descriptors}
\end{figure*}

\begin{figure*}
    \centering
    \includegraphics[page=9,width=\textwidth, clip,trim=0cm 48cm 2cm 66cm]{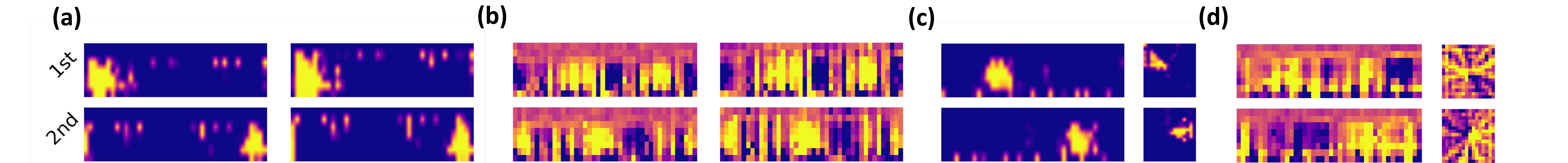}
    \caption{Comparison between the first two learned descriptors between the preliminary GeoDTR and the proposed \ourmodel{}. (a) and (c) are descriptors from the proposed \ourmodel{} trained with or without polar transformation respectively. (b) and (c) are descriptors from the preliminary GeoDTR trained with or without polar transformation respectively. In each sub-figure, the left is the ground image descriptor and the aerial image descriptor is on the right-hand side. Noticed that despite the polar transformation, \ourmodel{} learns more concentrated regions than GeoDTR which implies that \ourmodel{} learns a more distinguishable latent representation.}
    \label{fig:decriptor_compare}
\end{figure*}

\section{Discussion}
\label{sec::discussion}
\subsection{Discussion of Geometric Layout Descriptors}
Same-area performance has always been the main focus of Cross-View Geo-Localization (CVGL) methods for a long time.
While the increase in same-area performance gradually reaches its margin,
there is still a huge performance gap between cross-area and same-area cases.
In this work, we propose to highlight the cross-area performance as it reflects the actual utility of a CVGL model in real-world scenarios as discussed in \Cref{sec::introduction}.
Furthermore, the failure of generalizing to unseen areas might indicate overfitting of the model under consideration.

One possible way to address the cross-area generalization is to exploit the correspondence between the layouts of visual features in the ground and aerial images.
The geometric constraints underlying this correspondence are the same despite the dramatic change in the appearance of visual features from different areas.
Following this philosophy, we introduce a specialized pathway in \ourmodel{} for capturing the geometric layout of visual features.

Concretely, we model the geometric layout through a set of mask-like descriptors $\mathbf{q}$, which at output modulate the raw backbone feature (see~\Cref{eq:main}). 
Elements in a descriptor take value from the range $[0,1]$ and
each descriptor only deviates from zero at scattered areas that distribute across the whole spatial range.
Consequently, a descriptor filters out distinct ``active'' areas from the backbone feature for the final comparison. 
The essential characteristics of Geometric Layout Descriptors (GLD) can be expressed as follows,
\begin{itemize}
    \item Capturing the correlations among visual features, i.e., visual features within the active areas are considered to be correlated;
    \item Presenting the global property, i.e., the active areas distribute across the whole spatial range.
\end{itemize}
Our descriptor-based representation of geometric layout shares other benefits.
First, during the forward pass, the descriptors help pick up discriminative components from the raw backbone feature, leading to more efficient use of information in the backbone feature.
\ro{Second, only elements within the active areas receive significant gradient signals during back-propagation. This makes the model training more targeted to focus on discriminative features, resulting in the model easier to learn better latent representations.}

To justify the above picture and fully demonstrate the power of our model, 
we visualize the ground and aerial descriptors in~\Cref{fig:descriptors} for cases when \ourmodel{} is trained with polar transformed aerial images and normal aerial images (without polar transformation), respectively. As shown in the~\Cref{fig:descriptors}(a), we first note a strong alignment between descriptors of a given aerial-ground pair. It is clear to see that the corresponding descriptors share very similar salient values. More strikingly, such an alignment still exists when our model is trained with normal ground images. In~\Cref{fig:descriptors}(b), we unroll the descriptors for the normal aerial image by polar transformation. We observe that apart from the deformation brought by the polar transformation, the locations of salient patterns in the aerial descriptors match those in the corresponding ground descriptors. This indicates the ability of \ourmodel{} to grasp the geometric correspondence even without the guidance of polar transformation and  it also demonstrates the reason why the performance of \ourmodel{} training with or without polar transformation is extremely close and both achieve outstanding performance. Moreover, we compare the first two descriptors from the preliminary GeoDTR and the proposed \ourmodel{} trained with polar transformation and without it in~\Cref{fig:decriptor_compare}. \ro{It is clear to see that despite the polar transformation, \ourmodel{} learns more concentrated regions than GeoDTR which implies that fewer dimensions in the learned raw latent representations are needed to distinguish the aerial and ground views. By combining the overall performance improvement from GeoDTR to \ourmodel{}, it implies that \ourmodel{} learns a more distinguishable latent representation. 
}

\subsection{Discussion of Contrastive Hard Samples Generation}
Besides the geometric layout, 
we further integrate the proposed novel sampling strategy named Contrastive Hard Samples Generation (\chsg{}), which produces hard samples that share similar patterns while differing in layout in a single training batch.
This introduced intra-batch contrast over hard samples enables the \ourmodel{} to efficiently learn discriminative features. 
Ultimately, \chsg{} can push the model away from the local minimum and achieve better cross-area performance. 
Moreover, different from existing hard mining strategies in CVGL~\cite{revisiting,Vigor}, \chsg{} does not rely on a mining pool to maintain the global similarities among all the training samples. Thus, \chsg{} is more efficient in terms of both training time and memory usage.

\begin{figure}
    \centering
    \includegraphics[width=0.48\textwidth]{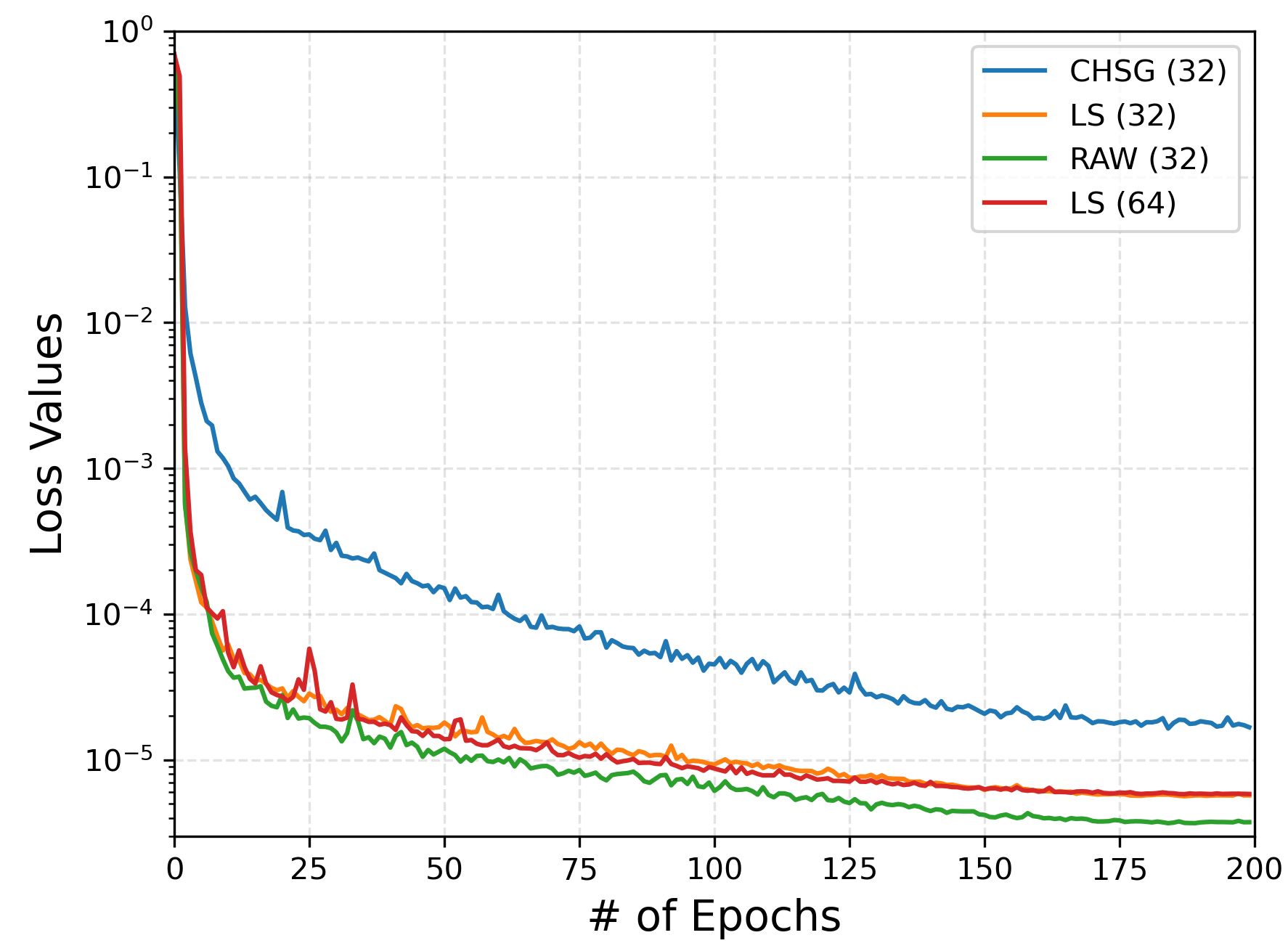}
    \caption{Comparison of triplet loss values of \ourmodel{} during training with different configurations on CVUSA dataset. ``\chsg{}'' stands for training with \chsg{}. ``LS'' stands for training with LS techniques. ``RAW'' stands for training without LS and \chsg{}. The number in the brackets is the batch size.}
    \label{fig:loss_comparison}
\end{figure}

\begin{figure}
    \centering
    \includegraphics[width=0.48\textwidth]{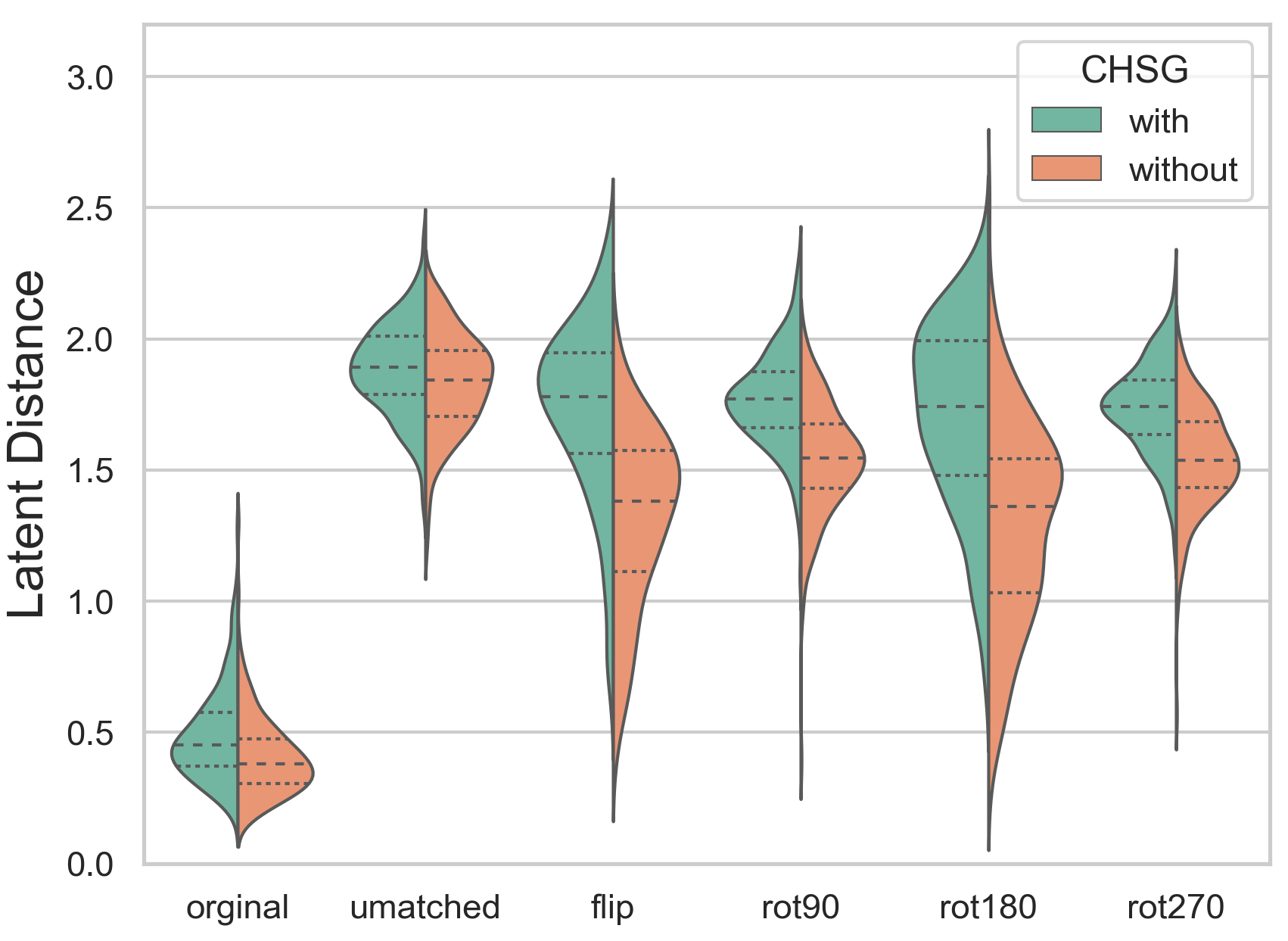}
    \caption{\ro{Visualization of the distributions of latent feature distances using violin plot of our \ourmodel{} training with \chsg{} and training without \chsg{} on $200$ random test aerial-ground pairs.} From left to right on the x-axis, ``original'' stands for original aerial-ground pairs, and ``unmatched'' stands for random unmatched aerial-ground pairs. ``flip'', ``rot90'', ``rot180'', ``rot270'' are fixing the aerial image, and flipping or rotating the ground images horizontally or in certain degrees respectively.}
    \label{fig:latent_feature_difference}
\end{figure}

\input{tables/CHSG-2}
Results in~\Cref{tab:chsg,tab:chsg_LS_batch} confirm the benefits of \chsg{} on our model, especially the improvement in cross-area performance.
In order to understand how the \chsg{} improves the model,
we first visualize the loss values during training under different configurations in~\Cref{fig:loss_comparison}.
In the plot, the experiment trained with vanilla triplet loss serves the baseline and is compared with the ones trained with LS and \chsg{}.
To demonstrate the effects of intra-batch contrast, two configurations of training with LS are included, one with a batch size of $32$ and the other of $64$.
While all models converge at a similar rate, applying \chsg{}
leads to a significantly and consistently higher loss than the other three configurations, thus providing a stronger supervision signal for the training.
The losses of training with LS are constantly higher than the one training without it but have a noticeable gap to the loss of training with \chsg{}. 
It is interesting to note that the loss curve of LS ($64$) is almost identical to that of LS ($32$). 
Recall that \chsg{} generates two variations from each original pair.
In this sense, it doubles the batch size, and therefore the difference between \chsg{} ($32$) and  LS ($64$) indicates that the increase in loss is a result of introducing hard contrastive samples. 

Next, we seek to investigate the different behaviors of the model trained with \chsg{} and with LS alone. 
Specifically, we care about the latent feature distance between a ground image and an aerial image. 
This experiment is carried out with $200$ randomly sampled test data from the CVUSA dataset, 
and \Cref{fig:latent_feature_difference} shows the distribution of latent feature distance for different types of aerial-ground pairs.
In the plot, ``original'' refers to the ground truth aerial-ground pairs. ``unmatched'' stands for randomly constructed unmatched aerial-ground pairs. 
``flip'', ``rot90'', ``rot180'' and ``rot270'' refers to those hard samples with respect to the ground truth aerial-ground pairs.
They are constructed by fixing the aerial image and flipping or rotating the ground truth ground image at certain degrees respectively. 
As expected, the ``original'' case has the lowest mean latent feature distance while the ``unmatched'' shows the largest. Noticeably the distributions are similar for models trained with and without \chsg{}.
However, those hard samples show different distributions of latent feature distance from the two models. 
While training without the proposed \chsg{}, the distance distributions of hard samples favor smaller values and, hence, the model cannot distinguish them well from the ground truth pairs.
On the other hand, when trained with \chsg{} the distance distributions of hard samples all shift towards larger values and become comparable with the ones of unmatched aerial-ground pairs.
This makes sense since flipping and rotation break the geometric correspondence in aerial-ground pairs. 
So that they can be considered as effective ``unmatched'' pairs. 
The ability to better distinguish hard samples endorses the outstanding performance of our model.

The previous discussion emphasizes the importance of the intra-batch contrast among hard samples.
It is interesting to have a fine-grained analysis of \chsg{} on how the semantic augmentation and layout simulation components contribute at the inter-batch and the intra-batch levels.
Thus, we compare four different settings of \chsg{} where the hard samples are generated by applying
\begin{itemize}
    \item different layout simulations only (L only)
    \item different semantic augmentations only (S only)
    \item same layout simulation and different semantic augmentations (same L $+$ S)
    \item same semantic augmentation and different layout simulations (same S $+$ L)
\end{itemize}
In the same L + S case, the effects of layout simulation happen at the inter-batch level while the effects of semantic augmentation at the intra-batch level (similar to the L + same S case).

\Cref{tab:chsg} summarizes the results. 
We found that inter-batch contrast with either semantic augmentation or layout simulation can boost the model on cross-area performance (L only or S only). 
Adding additional intra-batch contrastive signals (same L $+$ S or same S $+$ L) can further improve the accuracy. 
Best performance is achieved when both contrastive signals operate at the intra-batch level (L + S). Moreover, we found that adding either inter-batch or intra-batch contrast can hardly be harmful to the model on same-area performance. In other words, the results with adding contrast remain at the same level of performance as the one without adding any contrast (``Raw'' in \Cref{tab:chsg}). The above results confirm the contribution of the intra-batch contrast on the cross-area performance while not hurting the same-area results. 

%% file: tables/CHSG-2.tex
\begin{table}[!ht]
    \centering
    \caption{\label{tab:chsg} Comparison of \ourmodel{} trained with different configurations of the proposed \chsg{} on CVUSA dataset. ``RAW'' refers to training with original aerial-ground pairs. ``S'' is short for semantic augmentation. ``L'' is short for Layout simulation. ``same'' means the same operation (L or S) is applied to both contrastive samples $P_{L_{\gamma}}$ and $P_{L_{\delta}}$.}
    \begin{tabular}{ccccccc}
    \toprule 
    & Configuration & {\Rone{}} & {\Rfive{}} & {\Rten{}} & {\Ronep} \\\midrule
    \multirow{6}{*}{\rotatebox[origin=c]{90}{Same-area}}
    & Raw & 95.77\% & 98.93\% & 99.36\% & 99.81\% \\
    & S only & 94.78\% & 98.53\% & 99.12\% & 99.70\% \\
    & S + same L & 95.62\% & 99.00\% & 99.40\% & 99.84\% \\
    & L only & 95.52\% & 98.57\% & 99.14\% & 99.76\% \\
    & same S + L& 95.55\% & 98.64\% & 99.12\% & 99.75\% \\ 
    & L + S & 95.40\% & 98.44\% & 99.05\% &	99.75\% \\\midrule
    \multirow{6}{*}{\rotatebox[origin=c]{90}{Cross-area}}
    & Raw & 53.65\% & 74.91\% & 81.39\% & 93.27\% \\
    & S only & 55.77\% & 77.04\% & 82.28\% & 93.46\% \\
    & S + same L & 59.16\% & 79.72\% & 85.06\% & 94.55\% \\
    & L only & 58.09\% & 78.08\% & 84.18\% & 93.73\% \\
    & same S + L & 59.46\% & 79.82\% & 85.61\% & 94.46\% \\
    & L + S & 61.17\% & 80.22\% & 85.45\% & 94.56\% \\
    \bottomrule
    \end{tabular}
\end{table}

%% file: text/conclusion.tex
\section{conclusion}

In conclusion, we extend our preliminary work GeoDTR~\cite{GeoDTR} to address the problem of loss of information in geometric layout descriptor extraction and further explore the usage of LS techniques in improving cross-area cross-view geo-localization. In this paper, we propose \ourmodel{} to explicitly disentangles the geometric layout correlations from the raw features extracted by the backbone feature extractor via a novel Geometric Layout Extractor (GLE). The latent representation is a Frobenius product between the raw features and geometric layout extractors which effectively captures geometric correspondence between aerial and ground images, even without post-processing such as polar transformation. To push the limit of cross-area cross-view geo-localization, we take advantage of LS techniques and form a Contrastive Hard Sample Generation (\chsg{}) sampling strategy. \chsg{} aims to generate hard aerial-ground image pairs by varying the geometric layout and low-level details. Thus, there is no need to mine hard samples from a global pool which cost extra computational resources. Our experiment demonstrated that the proposed \ourmodel{} achieves state-of-the-art (SOTA) results on CVACT\_val~\cite{liu2019lending} which is 
one of the most challenging same-area benchmarks. Meanwhile, \ourmodel{} maintains a comparable same-area performance with other SOTA methods on CVUSA~\cite{CVUSA} and VIGOR~\cite{Vigor}. Most importantly, the proposed \ourmodel{} achieves SOTA performance on all cross-area benchmarks including, CVUSA~\cite{CVUSA}, CVACT~\cite{liu2019lending}, and VIGOR~\cite{Vigor} with a considerable improvement from the existing methods.

For future research, considering the ability of generalization of the proposed \ourmodel{}, it is worth investigating the performance of \ourmodel{} under different environments, for example, low-light scenarios or different seasons. It is also worth extending the usage of \chsg{} under other contrastive learning or image retrieval tasks to see how it works.

%% file: text/Appendix.tex
\section*{Appendix}
\label{sec::appendix}
\input{tables/VIGOR_ablation}

\begin{table*}[!ht]
    \centering
    \caption{\ro{Architecture comparison between the proposed \ourmodel{} and recently proposed SAIG~\cite{SAIG} on CVUSA~\cite{CVUSA} and CVACT~\cite{liu2019lending} datasets with same-area and cross-area evaluation. ``Same-area'' is training and testing on the same benchmark. ``Cross-area'' is training and testing on different benchmarks (i.e. CVUSA $\rightarrow$ CVACT or CVACT $\rightarrow$ CVUSA).}}
    \label{tab:archietecture_comparison}
    \begin{tabular}{cccccccccc}
    \toprule 
    \multirow{2}{*}{\ro{Training Set}} & \multirow{2}{*}{Method} & \multicolumn{4}{c}{Same-area} & \multicolumn{4}{c}{Cross-area} \\
    \cmidrule(lr){3-6}
    \cmidrule(lr){7-10}
     & & \Rone{} & \Rfive{} & \Rten{} & \Ronep  & \Rone{} & \Rfive{} & \Rten{} & \Ronep \\ \midrule
    \multirow{4}{*}{CVUSA} \
    & SAIG~\cite{SAIG} w/o SAM & 92.71\% & 97.92\% & 98.89\% & 99.71\% & 15.29\% & 33.07\% & 42.14\% & 72.95\% \\ 
    & \ourmodel{} w/o \chsg{} & 94.09\% & 98.68\% & 99.31\% & 99.82\% & 40.79\% & 62.53\% & 70.44\% & 89.14\% \\
    & SAIG~\cite{SAIG} w/ \chsg{} & 93.41\% & 99.08\% & 99.03\% & 99.73\% & 26.82\% & 47.47\% & 56.22\% & 81.95\% \\ 
    & \ourmodel{} w/ \chsg{} & 95.05\% & 98.42\% & 98.92\% & 99.77\% & 60.16\% & 79.97\% & 84.67\% & 94.48\% \\ \midrule
    \multirow{4}{*}{CVACT} 
    & SAIG~\cite{SAIG} w/o SAM & 84.42\% & 94.09\% & 95.57\% & 98.49\% & 18.97\% & 35.60\% & 44.28\% & 75.33\% \\
    & \ourmodel{} w/o \chsg{} & 87.49\% & 94.92\% & 96.26\% & 98.50\% & 29.15\% & 49.18\% & 58.41\% & 83.49\% \\
    & SAIG~\cite{SAIG} w/ \chsg{} & 85.41\% & 95.12\% & 96.33\% & 98.68\% & 32.49\% & 53.83\% & 62.85\% & 86.31\% \\ 
    & \ourmodel{} w/ \chsg{} & 87.76\% & 95.50\% & 96.50\% & 98.32\% & 52.56\% & 73.08\% & 79.82\% & 94.80\% \\
    \bottomrule
    \end{tabular}
\end{table*}

\subsection{Ablation study of \chsg{}, GLE, and backbone on VIGOR dataset}
\label{sec:ablation_vigor}
\common{To further demonstrate the effectiveness of our proposed new Geometric Layout Extractor (GLE), \chsg{}, and different backbones, we conducted another ablation study as shown in~\Cref{tab:VIGOR_ablation}. In~\Cref{tab:VIGOR_ablation}, we ablate the \chsg{}, geometric layout extractor, and backbone in our proposed \ourmodel{}. By comparing different GLEs with the same backbone and CHSG configurations, we can see that the new GLE improves the same-area performance. For example, while training with ResNet-34 backbone and without \chsg{}, R@1 increases from $56.51\%$ to $57.24\%$. Similarly, the new GLE improves R@1 from $55.14\%$ to $59.01\%$ with ConvNeXt-T backbone and \chsg{}. We also observed that \chsg{} significantly boosts the cross-area performance, especially on the ConvNeXt-T backbone (i.e. R@1 increases from $26.15\%$ to $36.01\%$ in the cross-area experiment). However, it is noted that the proposed new GLE can improve same-area performance while slightly decreasing cross-area performance when training without \chsg{} (for instance, training with ConvNeXt-T and without \chsg{} the R@1 decreases from $27.04\%$ to $26.15\%$). This slight decrease can be attributed to the new GLE that better captures the spatial correlations in the input images which might cause the model overfitting to them. Thus, we suggest applying both the proposed new GLE and the \chsg{} at the same time while training the model to obtain the best performance in both the same-area and cross-area experiments.}

\subsection{Architecture Comparison with SAIG}
\label{sec:arch_compare_saig}

\ro{To better compare the architecture of our \ourmodel{} to the recently proposed SAIG architecture which is specifically designed for cross-view image geo-localization, we experimented to compare them on CVUSA and CVACT datasets with same-area and cross-area evaluations. Different from experiments in~\cref{tab:GLE}, we did not employ polar transformation here for fair comparisons. To be noticed, we did not employ CHSG and SAM for our \ourmodel{} and SAIG, respectively, since our goal is to compare the architecture alone. To have a more comprehensive and fair comparison, we did not use polar transformation as pre-processing. The results are shown in~\Cref{tab:archietecture_comparison}. In this experiment, we ablate SAIG~\cite{SAIG} with SAM~\cite{SAM} technique and our \ourmodel{} with the proposed \chsg{}. As shown in this table, the architecture of our proposed \ourmodel{} is better than the recently proposed SAIG~\cite{SAIG} in both same-area and cross-area experiments while training without SAM~\cite{SAM} or \chsg{}. Specifically, in the cross-area experiment, our \ourmodel{} outperforms SAIG~\cite{SAIG} substantially, demonstrating the advantage of our GLE design that efficiently extracts geometric layout while avoiding overfitting to low-level details. While applying \chsg{}, we observe that both methods gain performance increase on same-area and cross-area experiments. To be noted that our proposed \ourmodel{} has more substantial improvements on the cross-area experiment. This might be attributed to the disentangling process we proposed in~\Cref{eq:main} that better generalizes on unseen data.}

\begin{table*}[!ht]
    \centering
    \scriptsize
    \caption{Experiment results by removing the sigmoid activation function at the output of the GLE on CVUSA and CVACT datasets. ``PT'' stands for polar transformation. Numbers in the brackets indicate the performance difference compared with our \ourmodel{} in~\Cref{tab:same_area}.}
    \label{tab:activation_ablation}
    \begin{tabular}{ccccccccc}
    \toprule 
     \multirow{2}{*}{PT} & \multicolumn{4}{c}{Same-area} & \multicolumn{4}{c}{Cross-area} \\
    \cmidrule(lr){2-5}
    \cmidrule(lr){6-9}
     & \Rone{} & \Rfive{} & \Rten{} & \Ronep  & \Rone{} & \Rfive{} & \Rten{} & \Ronep \\ [5pt]
    \multicolumn{9}{l}{\textit{on CVUSA}} \\ \midrule
    \checkmark & 93.97\% (-1.43\%) & 98.15\% (-0.29\%) & 98.89\% (-0.16\%) & 99.71\% (-0.11\%) & 59.04\% (-2.13\%) & 79.59\% (-0.63\%) & 85.06\% (-0.39\%) & 94.11\% (-0.45\%) \\ 
    $\times$ & 94.06\% (-0.99\%) & 98.35\% (-0.07\%) & 99.93\% (+0.01\%) & 99.71\% (-0.06\%) & 59.01\% (-1.15\%) & 78.87\% (-1.10\%) & 84.26\% (-0.41\%) & 94.15\% (-0.33\%) \\
    [5pt]
    \multicolumn{9}{l}{\textit{on CVACT}} \\
    \midrule
    \checkmark & 85.62\% (-1.99\%) & 94.57\% (-0.91\%) & 95.95\% (-0.57\%) & 98.06\% (-0.28\%) & 50.07\% (-3.82\%) & 73.13\% (-1.43\%) & 80.17\% (-0.93\%) & 94.15\% (-0.78\%) \\
    $\times$ & 85.53\% (-2.23\%) & 94.50\% (-1.00\%) & 95.71\% (0.79\%) & 98.11\% (-0.21\%) & 48.91\% (-3.65\%) & 69.77\% (-3.32\%) & 77.05\% (-2.77\%) & 94.27\% (-0.53\%) \\
    \bottomrule
    \end{tabular}
\end{table*}

\begin{table*}[!ht]
    \centering
    \scriptsize
    \caption{Experiment results by setting predicted GLDs to all-ones matrices during training on CVUSA and CVACT datasets. ``PT'' stands for polar transformation. Numbers in the brackets indicate the performance difference compared with our \ourmodel{} in~\Cref{tab:same_area}.}
    \label{tab:all1_ablation}
    \begin{tabular}{ccccccccc}
    \toprule 
     \multirow{2}{*}{PT} & \multicolumn{4}{c}{Same-area} & \multicolumn{4}{c}{Cross-area} \\
    \cmidrule(lr){2-5}
    \cmidrule(lr){6-9}
     & \Rone{} & \Rfive{} & \Rten{} & \Ronep  & \Rone{} & \Rfive{} & \Rten{} & \Ronep \\ [5pt]
    \multicolumn{9}{l}{\textit{on CVUSA}} \\ \midrule
    \checkmark & 92.81\% (-2.59\%) & 98.02\% (-0.42\%) & 98.91\% (-0.14\%) & 99.71\% (-0.04\%) & 52.55\% (-8.62\%) & 74.23\% (-5.99\%) & 80.62\% (-4.83\%) & 93.01\% (-1.55\%) \\ 
    $\times$ & 92.33\% (-2.72\%) & 97.90\% (-0.52\%) & 98.76\% (-0.16\%) & 99.71\% (-0.06\%) & 48.27\% (-11.89\%) & 71.94\% (-8.03\%) & 79.46\% (-5.21\%) & 93.23\% (-1.25\%) \\
    [5pt]
    \multicolumn{9}{l}{\textit{on CVACT}} \\
    \midrule
    \checkmark & 84.27\% (-3.34\%) & 94.25\% (-1.23\%) & 95.61\% (-0.91\%) & 98.27\% (-0.07\%) & 46.27\% (-7.62\%) & 68.45\% (-6.11\%) & 76.24\% (-4.86\%) & 93.75\% (-1.18\%) \\
    $\times$ & 83.33\% (-4.43\%) & 94.52\% (-0.98\%) & 96.05\% (0.45\%) & 98.39\% (+0.07\%) & 44.36\% (-8.20\%) & 66.81\% (-6.27\%) & 75.39\% (-4.43\%) & 94.36\% (-0.44\%) \\
    \bottomrule
    \end{tabular}
\end{table*}

\subsection{Further ablation study of GLDs}
\label{sec::ablation_GLD}
Geometric Layout Descriptors (GLDs) play an important role in capturing the layout information in the raw backbone features. In our model \ourmodel{}, GLDs are designed to be mask-like so that they reflect geometric information rather than low-level features. Specifically, we included a sigmoid activation at the end of the Geometric Layout Extractor (GLE) module, which maps each element in the output into the range between $0$ and $1$.
Here, we conduct two further ablation studies to demonstrate the effectiveness of the current implementation of GLDs. 

\subsubsection{Removing the Sigmoid activation function}
In the first experiment, we removed the sigmoid function from the GLE.
Consequently, elements of the output of GLE take values from the range $[-\infty, \infty]$.
\Cref{tab:activation_ablation} summarizes the performance of resulting models. 
Clearly, removing the sigmoid function decreases the performance of \ourmodel{} on both CVUSA and CVACT datasets in same-area and cross-area protocols, demonstrating the effectiveness of the design of our proposed GLE. Especially in cross-area results on CVACT dataset, we can observe a significant performance drop compared with the original model. On the CVUSA dataset, such performance drops are less drastic. This might be due to the fact that the CVACT dataset is densely sampled in a single city, hence being more challenging than the CVUSA dataset. \\
\subsubsection{Replacing GLDs with all-ones descriptors}
To further study the effectiveness of the proposed GLE, we replace the learned GLDs with dummy descriptors whose elements are fixed to be $1$ during both training and testing. Such all-ones descriptors evenly weight (modulate) all elements in the backbone feature, thereby imposing no constraint based on the geometric layout.
In this sense, the Geometric Layout pathway in \ourmodel{} is manually muted, and it breaks the disentanglement process.
The results are shown in~\Cref{tab:all1_ablation}. Significant performance drops can be observed across all the experiments in different datasets and protocols. More importantly, there are noticeable performance gaps that training with polar transformation is constantly better than training without it. This is in opposite to the results in~\Cref{tab:same_area}, where \ourmodel{} displays similar performance regardless of whether training with or without polar transformation. The above results in~\Cref{tab:all1_ablation} indicate that 
the ability to capture the geometric layout information is a crucial component for boosting the cross-view geo-localization performance in our model.

\begin{table*}[!ht]
    \centering
    \scriptsize
    \caption{Ablation study of the counterfactual training paradigm on CVUSA and CVACT datasets. ``PT'' stands for polar transformation. Numbers in the brackets indicate the performance difference compared with our \ourmodel{} in~\Cref{tab:same_area}.}
    \label{tab:CF_ablation}
    \begin{tabular}{ccccccccc}
    \toprule 
     \multirow{2}{*}{PT} & \multicolumn{4}{c}{Same-area} & \multicolumn{4}{c}{Cross-area} \\
    \cmidrule(lr){2-5}
    \cmidrule(lr){6-9}
     & \Rone{} & \Rfive{} & \Rten{} & \Ronep  & \Rone{} & \Rfive{} & \Rten{} & \Ronep \\ [5pt]
    \multicolumn{9}{l}{\textit{on CVUSA}} \\ \midrule
    \checkmark & 95.10\% (-0.30\%) & 98.40\% (-0.04\%) & 98.97\% (-0.08\%) & 99.76\% (+0.01\%) & 59.15\% (-2.02\%) & 78.58\% (-1.64\%) & 84.39\% (-1.06\%) & 94.03\% (-0.53\%) \\ 
    $\times$ & 94.04\% (-1.01\%) & 98.19\% (-0.23\%) & 98.93\% (+0.01\%) & 99.76\% (-0.01\%) & 57.76\% (-2.40\%) & 79.01\% (-0.96\%) & 84.50\% (-0.17\%) & 94.24\% (-0.24\%) \\
    [5pt]
    \multicolumn{9}{l}{\textit{on CVACT}} \\
    \midrule
    \checkmark & 86.04\% (-1.57\%) & 95.01\% (-0.47\%) & 96.14\% (-0.38\%) & 98.20\% (-0.14\%) & 49.73\% (-4.16\%) & 72.14\% (-2.42\%) & 78.81\% (-2.29\%) & 93.92\% (-1.01\%) \\
    $\times$ & 86.50\% (-1.26\%) & 95.01\% (-0.49\%) & 96.26\% (-0.24\%) & 98.33\% (+0.01\%) & 49.48\% (-3.08\%) & 71.01\% (-2.07\%) & 78.63\% (-1.19\%) & 94.16\% (-0.64\%) \\
    \bottomrule
    \end{tabular}
\end{table*}

\subsection{Ablation study of CF learning schema}
\label{sec::ablation_CF}
The counterfactual (CF) learning schema was originally proposed in our preliminary work~\cite{GeoDTR}. Here, we conduct an ablation study to explicitly demonstrate the effectiveness of CF in the current model \ourmodel{}. 
The experiments are carried out on the CVUSA and CVACT datasets with same-area and cross-area protocols. The results are shown in~\Cref{tab:CF_ablation}, which show similar pattern as in our previous work~\cite{GeoDTR}. First, we find that CF learning schema significantly increases performance on cross-area protocol, especially on the CVACT dataset. Secondly, we observe stronger performance gain on the CVACT dataset than on the CVUSA dataset. This is more evident for the same-area protocol, and we believe that this could be attributed to the saturation of the CVUSA dataset. In conclusion, the ablation study demonstrates that the proposed CF learning schema is able to improve the model performance under both same-area and cross-area protocols.

%% file: tables/VIGOR_ablation.tex
\begin{table*}[!ht]
    \centering
    \caption{\label{tab:VIGOR_ablation} \common{Ablation study of the proposed Geometric Layout Extractor (GLE) and \chsg{} on VIGOR dataset. In the GLE configuration, ``v1'' stands for the GLE from our preliminary work~\cite{GeoDTR}. ``v2'' stands for the GLE proposed in this work.}}
    \begin{tabular}{cc@{\hspace*{2mm}}c@{\hspace*{2mm}}cccccccccc}
    \toprule 
    \multicolumn{3}{c}{Configurations} & \multicolumn{5}{c}{Same-area} & \multicolumn{5}{c}{Cross-area} \\
    \cmidrule(lr){1-3}
    \cmidrule(lr){4-8}
    \cmidrule(lr){9-13}
     \chsg{} & GLE & Backbone & \Rone{} & \Rfive{} & \Rten{} & \Ronep & Hit Rate & \Rone{} & \Rfive{} & \Rten{} & \Ronep & Hit Rate \\ \midrule
    $\times$ & v1 & ResNet-34 & 56.51\% & 80.37\% & 86.21\% & 99.25\% & 61.76\% & 30.02\% & 52.67\% & 61.45\% & 94.40\% & 30.19\% \\
    $\checkmark$ & v1 & ResNet-34 & 55.53\% & 80.05\% & 86.31\% & 99.41\% & 62.09\% & 32.57\% & 54.97\% & 63.44\% & 93.86\% & 35.47\% \\
    $\times$ & v1 & ConvNeXt-T & 56.63\% & 80.85\% & 86.96\% & 99.48\% & 62.79\% & 27.04\% & 48.85\% & 57.61\% & 93.96\% & 29.43\% \\
    $\checkmark$ & v1 & ConvNeXt-T & 55.14\% & 79.30\% & 85.73\% & 99.23\% & 60.90\% & 33.86\% & 55.93\% & 63.43\% & 94.22\% & 36.54\% \\ 
    $\times$ & v2 & ResNet-34 & 57.24\% & 81.89\% & 87.76\% & 99.50\% & 63.66\% & 29.95\% & 50.94\% & 59.75\% & 93.04\% & 31.82\% \\
    $\checkmark$ & v2 & ResNet-34 & 58.24\% & 82.27\% & 88.04\% & 99.54\% & 64.79\% & 34.14\% & 57.58\% & 66.04\% & 95.29\% & 37.97\% \\
    $\times$ & v2 & ConvNeXt-T & 58.46\% & 83.85\% & 89.36\% & 99.61\% & 65.25\% & 26.15\% & 49.39\% & 59.06\% & 94.58\% & 29.80\% \\
    $\checkmark$ & v2 & ConvNeXt-T & 59.01\% & 81.77\% & 87.10\% & 99.07\% & 67.41\% & 36.01\% & 59.06\% & 67.22\% & 94.95\% & 39.40\% \\
    \bottomrule
    \end{tabular}
\end{table*}